\documentclass[journal,twoside]{IEEEtran}
\usepackage{cite}
\usepackage{hyperref}

\ifCLASSINFOpdf
\else
\fi

\usepackage{amsmath}
\usepackage{algorithm}
\usepackage{algpseudocode}
\usepackage{array}
\ifCLASSOPTIONcompsoc
  \usepackage[caption=false,font=normalsize,labelfont=sf,textfont=sf]{subfig}
\else
  \usepackage[caption=false,font=footnotesize]{subfig}
\fi

\usepackage{url}
\usepackage{todonotes}
\usepackage{tikz}
\usetikzlibrary{shapes,arrows}
\usetikzlibrary{matrix,chains,shapes.geometric,positioning,decorations.pathreplacing,arrows}
\usetikzlibrary{arrows,calc,decorations.markings,math,arrows.meta}
\usepackage{multirow}
\usepackage{siunitx}

\usepackage{cleveref}

\usepackage{mlmacros}		%
\variables{x,z,y,u,r}
\variables[mean,std]{\mu,\sigma}
\varmacros{E}{\mathcal{E}}
\probdists{p,q,r}
\probdists[policydist]{\pi}
\probdists[policy]{\policydist[\theta]{u_t}{z_t}}
\probdists[reward]{\r{z_t, u_{t}}}
\probdists[apxreward]{\r[\xi]{z_t, u_{t}}}
\probdists[apxnextreward]{\r[\xi]{z_{t+1}, u_{t+1}}}
\probdists[transition]{\p{z_{t+1}}{z_t, u_t}}
\probdists[apxtransition]{\p[\xi]{z_{t+1}}{z_t, u_t}}
\probdists[apxvalue]{V^\pi_\phi(z_t)}
\probdists[apxvaluenext]{V^\pi_\phi(z_{t+1})}
\probdists[apxvaluenextN]{V^\pi_\phi(z_{t+H})}
\probdists[apxvaluenextNtarget]{V^\pi_{\phi'}(z_{t+H})}
\MkProbDist{P}{\mathbb{P}}

\usepackage{mlmacros}		%
\variables{x,z,u,s,A,B}
\variables[mean,std]{\mu,\sigma}
\varmacros{E}{\mathcal{E}}

\newcommand{\ztran}{\p[\xi]{z_t}{z\tm, s_t, u\tm}}
\newcommand{\stran}{\p[\xi]{s_t}{s\tm, z\tm, u\tm}}
\newcommand{\qmeas}{\q[\mathrm{meas}]{z_t}{x_t}}
\newcommand{\qtran}{\q[\mathrm{trans}]{z_t}{z\tm,s_t,u\tm}}

\newcommand{\specialcell}[2][c]{%
  \begin{tabular}[#1]{@{}c@{}}#2\end{tabular}}

\usepackage{xcolor}

\PassOptionsToPackage{printonlyused}{acronym}
  \usepackage{acronym} 

\newacro{mbrl}[MBRL]{model-based reinforcement learning}

\begin{document}

\title{Learning to Fly via \\Deep Model-Based Reinforcement Learning}

\author{Philip~Becker-Ehmck*,
        Maximilian~Karl*,
        Jan~Peters,
        and~Patrick~van~der~Smagt%
\thanks{* Authors have contributed equally.}%
\thanks{P. Becker-Ehmck, M. Karl and P. van der Smagt are with the Volkswagen Group, Machine Learning Research Lab, Munich, Germany (e-mail: philip.becker-ehmck@argmax.ai; karlma@argmax.ai).}%
\thanks{P. Becker-Ehmck and J. Peters are with Technical University of Darmstadt, Intelligent Autonomous Systems, Darmstadt, Germany.}%
\thanks{J. Peters is with the Max Planck Institute for Intelligent Systems, T{\"u}bingen, Germany.}%
}
\markboth{}%
{Becker-Ehmck \textit{\MakeLowercase{et al.}}: Learning to Fly via Deep Model-Based Reinforcement Learning}

\maketitle

\begin{abstract}
Learning to control robots without requiring engineered models has been a long-term goal, promising diverse and novel applications.
Yet, reinforcement learning has only achieved limited impact on real-time robot control due to its high demand of real-world interactions.
In this work, by leveraging a learnt probabilistic model of drone dynamics, we learn a thrust-attitude controller for a quadrotor through model-based reinforcement learning.
No prior knowledge of the flight dynamics is assumed; instead, a sequential latent variable model, used generatively and as an online filter, is learnt from raw sensory input.
The controller and value function are optimised entirely by propagating stochastic analytic gradients through generated latent trajectories. 
We show that ``learning to fly'' can be achieved with less than 30 minutes of experience with a single drone, and can be deployed solely using onboard computational resources and sensors, on a self-built drone.
\end{abstract}


\IEEEpeerreviewmaketitle

\section{Introduction}
\IEEEPARstart{D}{esigning} controllers for robots requires years of expertise and effort, fine tuning modules for state estimation, model-predictive control and contact scheduling.
As a final product, one receives a controller that is typically only applicable to one specific robot.
In contrast, deep reinforcement learning (RL) presumes no prior knowledge of a robots dynamics and promises broad applicability.
However, applying deep RL to robotic systems faces many challenges, among them poor sample efficiency.
This shortcoming is often circumvented by finding ways to parallelise the learning over many agents (\ie many identical robotic setups) and by automatic resetting of the environment~\cite{gu2017deep,kalashnikov2018qt}.
As an alternative to expensive real-world rollouts, simulators have commonly been employed, allowing us to generate more data without actually operating the robot.
This approach has been the most successful way to deploy learnt controllers on real hardware~\cite{peng2018sim, andrychowicz2018learning, tan2018sim, chebotar2019closing, lin2019flying}.
However, this just avoids the underlying issue that we wanted to address:
while controller learning is now generic, all the expert knowledge is still required to build the simulator.
Thus, a generic solution using no to little prior knowledge requires us to also learn the simulator.

All of these restrictions become obvious when working on drones.
Simulating the dynamics of a drone, esp.\ a self-built one, is complex.
Flying many flights with a drone is difficult because of battery change by human operators; flying many drones in parallel is difficult because of space constraints.
We therefore focus on an approach that is more data-efficient as we have to reduce data collection needs to a minimum.
We learn a full dynamics simulation of the drone, employing neural variational inference methods to train a latent state-space model from observational data~\cite{krishnan2017structured, karl2017deep, fraccaro2017disentangled, becker2019switching}.
This generic framework is theoretically applicable to all kinds of dynamical systems as it is agnostic to sensorimotor configuration.

\begin{figure}[t]
  \centering%
  \includegraphics[width=\columnwidth]{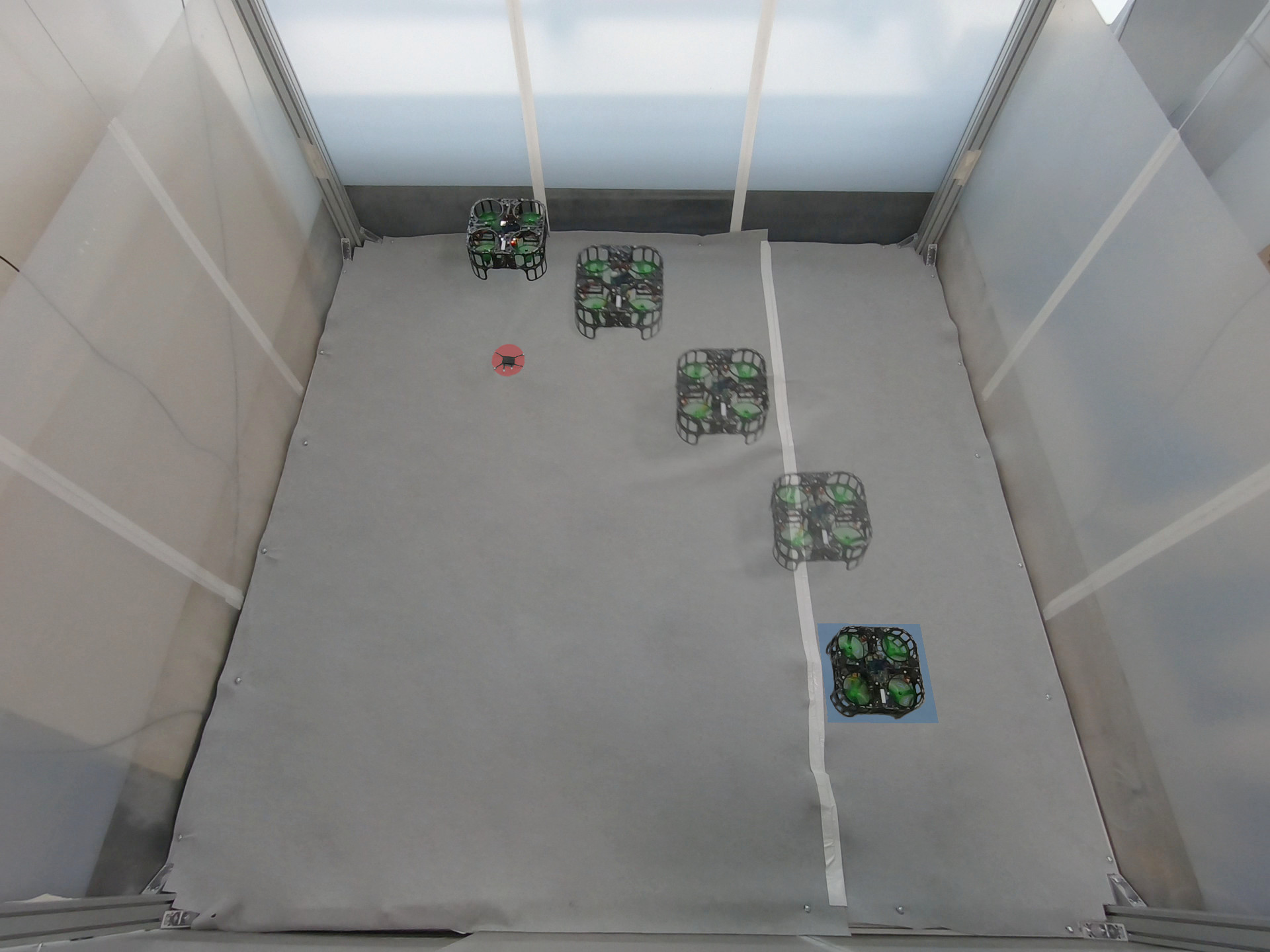}%
  \caption{A learnt controller flying a quadrotor to a goal marker. The controller was optimised using rollouts from a learnt dynamics model and is executed onboard. Video: \url{https://youtu.be/e5buJL_DYgA}.}
  \label{fig:drone-trajectory-no-yaw}
\end{figure}

As model bias represents a huge challenge of this approach, most prior work on learning controllers for quadrotors has been conducted using a predefined drone dynamics model, \eg~\cite{hwangbo2017control, lin2019flying, koch2019neuroflight}.
Typically, they employ either imitation learning or model-free RL algorithms where the predefined simulator is used as a data generator.
As a drawback, the simulator's internal structures can usually not be exploited for optimisation and its gap to reality might be high.
We will argue that using model derivatives of a learnt and differentiable simulator is a crucial advantage over model-free approaches.
The learning of simulators from data may also help reduce the simulation-to-reality gap.
Further, with the predefined simulator being defined in an engineered state-space, the learnt controllers often also act on full state observations.
Without the need for building an internal model of the world, they are inapplicable to partially observed environments.\\

Our principal contribution is the learning of a controller capable of flying a self-built drone to marked positions (see \Cref{fig:drone-trajectory-no-yaw}) using \ac{mbrl} methods.
The controller is entirely optimised in a learnt drone dynamics model from raw sensor data and can deal with noisy and partial observations of its current state.
Both the learnt model---used for online state estimation---and controller are executed in real-time on the drone itself using only embedded computational resources (a Raspberry Pi 4).
By learning a model, we can be much more data efficient than model-free methods, requiring only 30 minutes (equivalent to approximately 25.000 model steps) of real-world flight on a single drone to find our final policy.
The controller acts on the same level a human would operate a quadrotor, providing roll, pitch, yaw and throttle control inputs.
It is supported by a low-level controller which keeps the drone stable in absence of control inputs---a feature which mainly simplifies the initial exploration policy.

During exploration, the drone flies into obstacles every now and then. 
We therefore built our own drone platform, enhanced with a robust frame, and mounted with various sensors.
Details of the hardware of the drone are explained in this paper.

\section{background}
\subsection{Notation \& Naming Conventions}
Throughout the paper, we consider observations $x_t$ and control inputs $u_t$ that form a trajectory or episode $\tau = (x_1, u_1, x_2, u_2, ..., u_{T-1}, x_T)$.
We denote a sequence of variables as $\xseq = x_{1:T} = (x_1, x_2, ..., x_T)$.

\subsection{Backpropagation through Stochastic Variables}
Computing $\nabla_{\theta} \expc[\q[\theta]{z}{\cdot}]{f(z)}$, i.e. a gradient \wrt to some parameters $\theta$ where the expectation is taken with respect to a distribution parametrised by $\theta$ is key for learning stochastic representations by gradient descent.
Function $f$ is assumed to be integrable and smooth.
One way to approximate this gradient is by employing the score function estimator (also called REINFORCE~\cite{williams1992simple} or likelihood ratio method~\cite{glynn1990likelihood}):
\begin{equation}
\label{eq:score-function-esitmator}
\nabla_{\theta} \expc[\q[\theta]{z}{\cdot}]{f(z)} = \expc[\q[\theta]{z}{\cdot}]{f(z)\nabla_{\theta}\log \q[\theta]{z}{\cdot}}.
\end{equation}
In practice, this estimator often suffers from high variance.
In this work, we instead make use of a pathwise Monte Carlo gradient estimator, a technique repopularised by~\cite{kingma2014auto, rezende2014stochastic}.
Instead of sampling from $z$ directly, one can sample from an auxiliary noise variable $\epsilon$ that is independent from the parameters $\theta$ and transform this into a sample of the original distribution:
\begin{equation}
\label{eq:reparameterization}
z = g_\theta(\epsilon) \quad \text{with} \quad \epsilon \sim p(\epsilon).
\end{equation}
Such a transformation $g$ exists for many continuous distributions and is in general required to be smooth and invertible.
This allows us to compute Monte Carlo estimates of expectations by sampling from $p(\epsilon)$ instead:
\begin{equation}
\label{eq:reparameterization-expectation-estimation}
\mathbb{E}_{q_{\theta}(z|x)}[f(z)] = \mathbb{E}_{p(\epsilon)}[f(g_\theta(\epsilon))].
\end{equation}
In this form, the gradient operator $\nabla_{\theta}$ can trivially be moved inside the expectation.
We made use of this technique, often called \textit{reparameterisation trick}, both for learning a latent variable model and for optimising a stochastic policy.

\subsection{Variational Autoencoder}
Let us consider some data set $X = \lbrace x^{(i)} \rbrace_{i=1}^N$ that we assume to be generated by a random process based on latent variables $z$.
Formally, this generative process can be expressed by
\begin{equation}
\p{x} = \int \p{x,z} \dint z = \int \p{x}{z}\p{z} \dint z .
\label{eq:vae-graphical-model}
\end{equation} 
When trying to perform inference, \ie finding the posterior probability $\p{z}{x}$, we are restricted to approximate methods if the generative model $\p{x}{z}$ is nonlinear, \eg a neural network with one hidden layer and nonlinearity.
\cite{kingma2014auto} introduced the \textit{Variational Autoencoder} (VAE), a neural variational inference method where the posterior $\p{z}{x}$ is approximated by a neural network $\q[\psi]{z}{x}$.
The model is optimised by maximising the \textit{Evidence Lower Bound} (ELBO) of the data log-likelihood:
\begin{equation}
\label{eq:vae-elbo}
\begin{aligned}
&\log \p{x} \geq \loss[\mathrm{ELBO}]{x; \xi, \psi}\\
&= \expc[\q[\psi]{z}{x}]{\ln \p[\xi]{x}{z}} - \kl{\q[\psi]{z}{x}}{\p{z}}.
\end{aligned}
\end{equation}
Using the previously described reparameterisation trick this can be optimised end-to-end by taking a single sample for the expectation.
This framework can be extended to time series models in various ways~\cite{krishnan2017structured, fraccaro2017disentangled, karl2017deep, becker2019switching}.
We focus on one particular extension that takes the form of latent state-space models.

\subsection{Latent State-Space Models}
Latent state-space models (LSSMs) are characterised by the following equations 
\begin{align}
\label{eq:ssm-transition}
z_t &= f(z_{t-1}, u_{t-1}, q_{t-1}) &q_{t-1} \sim p(\cdot)\\
\label{eq:ssm-emission}
x_t &= h(z_t, w_t) &w_t \sim p(\cdot)
\end{align}
where transition $f$ and emission $h$ are nonlinear functions with arbitrary noise distributions $q$ and $w$.
If transition $f$ and emission $h$ are linear and known, $w$ and $q$ are white noise and the state-space is Gaussian, we can do state estimation analytically using a Kalman filter/smoother~\cite{sarkka2013bayesian}.
In this work, we are interested in the case where both functions are nonlinear and unknown.
We would like to learn both a latent state-space representation as well as the latent dynamics simultaneously.
Sequential VAEs are one general method that allow us to do this.
Using the state space model assumptions that the transition and emission are Markovian, such a model can be optimised using a time-factorised ELBO\cite{krishnan2017structured, karl2017deep}:
\begin{align}
\label{eq:sequential-elbo}
\begin{split}
\p{\xseq}{\useq} \geq \; & \sum_{t=1}^{T}\expc[z_t \sim \q[\psi]{z_t}{\xseq, \useq}]{\log \p[\xi]{x_t}{z_t}} \\
&- \kl{\q[\psi]{z_t}{\xseq, \useq}}{\p[\xi]{z_t}{z_{t-1}, u_{t-1}}}.
\end{split}
\end{align}

\subsection{Reinforcement Learning}
\label{sec:background-rl}
We consider the regular Markov decision process (MDP)~\cite{bellman1957markovian} where both state space $\mathcal{Z}$ and action space $\mathcal{U}$ are continuous.
An agent starts in an initial state $z_0 \sim \p{z_0}$.
At every time $t$, the agent samples a control $u_t$ from its policy $\policy$ and transitions to a new state $z_{t+1}$ according to the dynamics $\transition$.
The agent receives a bounded reward $r_t \sim \reward$ reinforcing or punishing the behaviour.
The reward-to-go $R_t = \sum_{k=0}^\infty \gamma^k r_{t+k}$ is the accumulated discounted reward from now on, with a discount factor $\gamma \in [0;1)$.
Reinforcement Learning seeks to learn a policy that maximises the expected reward-to-go
\begin{equation}
J(\pi) = \expc[\tau \sim \p[\pi]{\tau}]{R_t},
\label{eq:mc-policy-objective}
\end{equation}
where $\p[\pi]{\tau}$ represents the distribution over trajectories induced by following policy $\pi$.
Optimising directly \wrt \Cref{eq:mc-policy-objective}, as is often done in policy search methods~\cite{deisenroth2013survey}, can suffer from high variance since we are using a single sampled trajectory to evaluate the performance of our policy.
Value-based methods introduce state or state-action value estimators which can reduce variance while introducing a bias.
The value of a state $z$ under policy $\pi$ is defined as the expected reward for following the policy from state $z$, i.e. $V^\pi(z) = \expc[\tau \sim \p[\pi]{\tau}]{R_t}{z_t = z}$.
The state-action value is defined as $Q^\pi(z_t,u_t) = r(z_t,u_t) + \gamma \expc[z_{t+1} \sim \transition]{V^\pi(z_{t+1})}$.
To scale to real-world problems, state and state-action value functions are typically represented by function approximators such as neural networks.
Actor-Critic methods are a hybrid approach that learn both a policy (called \textit{actor}) and value function (called \textit{critic}) that bootstrap each other.

When working in continuous action spaces, optimization is most commonly done using policy gradients where the policy parameters $\theta$ are updated in direction of the performance gradient~\cite{sutton2000policy}:
\begin{equation}
\expc[\tau \sim \p[\pi]{\tau}]{\nabla_\theta \log \policy Q^\pi(z_t,u_t)}.
\end{equation}
Notably, the policy gradient does not depend on the gradient of the state distribution.

Model-based reinforcement learning (MBRL) distinguishes itself from model-free methods by exploiting a (possibly) learnt dynamics model $\transition$ to optimise a policy. 
There are numerous ways of doing so.
In Dyna~\cite{sutton1991dyna} and derivative work, the learnt dynamics model is used exclusively as a data generator for optimising the policy using standard model-free algorithms.
Real-world data is here only used to fit the dynamics model.
Another category of algorithms uses model derivatives either for policy search or improved value estimation\cite{heess2015learning, feinberg2018model, byravan2019imagined}.

\section{Variational Latent Dynamics}
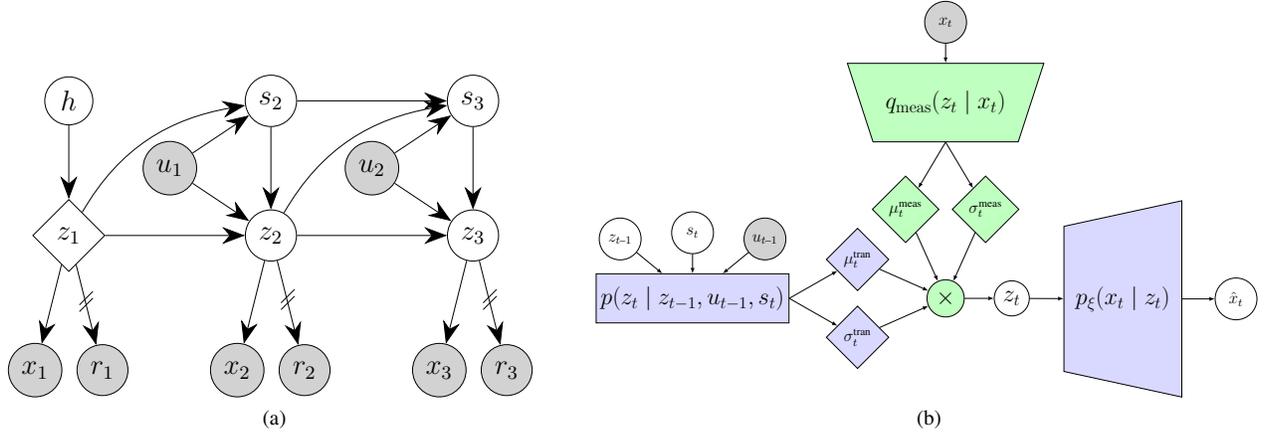
\begin{figure*}[t]
\centering
\null\hfill
\subfloat[]{\resizebox{0.8\columnwidth}{!}{\begin{tikzpicture}[
>={Stealth[scale=2.0]},
var/.style={
draw,
fill=white,
circle, 
minimum width={width("$z_{t+1}$")+2pt},
minimum height={height("$z_{t+1}$")+2pt},
font=\large},
input/.style={
draw,
fill=lightgray!70,
circle, 
minimum width={width("$z_{t+1}$")+2pt},
minimum height={height("$z_{t+1}$")+2pt},
font=\large},
det/.style={
draw,
fill=white,
diamond, 
font=\large},
network/.style={
draw,
fill=Aquamarine,
rectangle, 
minimum width={width("$z_{t+1}$")+4pt},
minimum height={height("$z_{t+1}$")+4pt},
font=\large}]

\def\tzero{3}
\def\tone{6}
\def\ttwo{9}

\node[var, draw=black!100] at (\tzero, 6) (h) {$h$};
\node[var, draw=black!100] at (\tone, 6) (s_2) {$s_2$};
\node[var, draw=black!100] at (\ttwo, 6) (s_3) {$s_3$};

\node[det, draw=black!100] at (\tzero, 4) (z_1) {$z_1$};
\node[var, draw=black!100] at (\tone, 4) (z_2) {$z_2$};
\node[var, draw=black!100] at (\ttwo, 4) (z_3) {$z_3$};

\node[input, draw=black!100] at (\tzero +1.5, 5) (u_1) {$u_1$};
\node[input, draw=black!100] at (\tone + 1.5, 5) (u_2) {$u_2$};

\node[input, draw=black!100] at (\tzero-0.5, 2) (x_1) {$x_1$};
\node[input, draw=black!100] at (\tone-0.5, 2) (x_2) {$x_2$};
\node[input, draw=black!100] at (\ttwo-0.5, 2) (x_3) {$x_3$};
\node[input, draw=black!100] at (\tzero+0.5, 2) (r_1) {$r_1$};
\node[input, draw=black!100] at (\tone+0.5, 2) (r_2) {$r_2$};
\node[input, draw=black!100] at (\ttwo+0.5, 2) (r_3) {$r_3$};

\node at (6, 2.5) (F) {};
\coordinate (f_1) at ([xshift=1.cm]F.east);

\draw (h) edge[->] (z_1);
\draw (z_1) edge[->] (x_1);
\draw (z_2) edge[->] (x_2);
\draw (z_3) edge[->] (x_3);
\draw (z_1) edge[->] coordinate (mr1) (r_1);
\draw (z_2) edge[->] coordinate (mr2) (r_2);
\draw (z_3) edge[->] coordinate (mr3) (r_3);

\draw[black,shift={(mr1)}](-0.1,-0.1)--(0.1,+0.1);
\draw[black,shift={(mr1)}](-0.1,+0.0)--(0.1,+0.2);
\draw[black,shift={(mr2)}](-0.1,-0.1)--(0.1,+0.1);
\draw[black,shift={(mr2)}](-0.1,+0.0)--(0.1,+0.2);
\draw[black,shift={(mr3)}](-0.1,-0.1)--(0.1,+0.1);
\draw[black,shift={(mr3)}](-0.1,+0.0)--(0.1,+0.2);

\draw (z_1) edge[->] (z_2);
\draw[->,out=60,in=190] (z_1) to (s_2);
\draw (z_2) edge[->] (z_3);
\draw[->,out=60,in=190] (z_2) to (s_3);

\draw (u_1) edge[->] (z_2);
\draw (u_2) edge[->] (z_3);
\draw (u_1) edge[->] (s_2);
\draw (u_2) edge[->] (s_3);

\draw (s_2) edge[->] (s_3);
\draw (s_2) edge[->] (z_2);
\draw (s_3) edge[->] (z_3);

\end{tikzpicture}}%
\label{fig:generative}}
\hfill
\subfloat[]{\resizebox{1.0\columnwidth}{!}{\begin{tikzpicture}[
var/.style={
draw,
fill=white,
circle, 
minimum width={width("$z_{t+1}$")+12pt},
minimum height={height("$z_{t+1}$")+12pt},
font=\large},
input/.style={
draw,
fill=lightgray!70,
circle, 
minimum width={width("$z_{t+1}$")+12pt},
minimum height={height("$z_{t+1}$")+12pt},
font=\large},
det/.style={
draw,
fill=white,
diamond, 
font=\large}]
\tikzset{
  font={\fontsize{18pt}{20}\selectfont}}
\tikzset{trapezium stretches=true}
\tikzset{>=latex}

\node [trapezium, trapezium angle=100, minimum width=5cm, minimum height=2cm, draw, fill=green!25] (rec) at (0,0) {};
\node at (rec.center) {$\q[\mathrm{meas}]{z_t}{x_{t}}$};

\node [below=3.5cm of rec, circle, draw, fill=green!25] (mult) {$\times$};
\node[rectangle, draw, fill=blue!15, left=3.5cm of mult, align=center, minimum height=1.25cm] (tra) {$\p{z_t}{z_{t-1},u_{t-1},s_t} $};
\node[circle, draw, right=.75cm of mult] (filter) {$z_t$};

\node [trapezium, fill=blue!15,  rotate=90, trapezium angle=80, minimum width=5cm, minimum height=3cm, draw, below right=1.2cm and 1cm of filter] (gen) {};
\node at (gen.center) {$\p[\xi]{x_t}{z_t}$};

\node [var, above left=0.5cm and -1cm of tra] (prev) {$z\tm$};
\node [var, above=0.5cm of tra] (s) {$s_t$};
\node [input, above right=0.5cm and -1cm of tra] (con) {$u\tm$};

\node [input, draw, above=.5cm of rec] (obs) {$x_{t}$};
\node [var, draw, below right=1.8cm and 2.5cm of gen.east] (obs2) {$\hat{x}_{t}$};

\node [det, fill=blue!15, above left=0.25cm and 1.5cm of mult] (mu_tran) {$\mu_t^{\textrm{tran}}$};
\node [det, fill=blue!15, below left=0.25cm and 1.5cm of mult] (sigma_tran) {$\sigma_t^{\textrm{tran}}$};
\node [det, above left=1.5cm and 0.25cm of mult, fill=green!25] (mu_meas) {$\mu_t^{\textrm{meas}}$};
\node [det, above right=1.5cm and 0.25cm of mult, fill=green!25] (sigma_meas) {$\sigma_t^{\textrm{meas}}$};

\path 
(rec.south) edge[->] (mu_meas)
(rec.south) edge[->] (sigma_meas)
(mu_meas) edge[->] (mult)
(sigma_meas) edge[->] (mult)
(filter) edge[->]  (gen)
(prev) edge[->]  (tra)
(con) edge[->] (tra)
(s) edge[->] (tra)
(tra.east) edge[->]  (mu_tran)
(tra.east) edge[->]  (sigma_tran)
(mu_tran) edge[->]  (mult)
(sigma_tran) edge[->]  (mult)
(mult) edge[->]  (filter)

(obs) edge[->]  (rec)
(gen) edge[->]  (obs2);


\end{tikzpicture}}%
\label{fig:fusion}}
\hfill\null
\caption{(a) Depicts the generative model consisting of latent states $z_t$ and $s_t$, observations $x_t$ and control inputs $u_t$. The crossed out arrow from latent state $z_t$ to reward $r_t$ indicates that the gradient is stopped. (b) Shows schematically how we combine the transition with the inverse measurement model in the inference network. Generative model is coloured blue, inference model green.}
\label{fig:sequence-model}
\end{figure*}

In this paper, we employ a variational LSSM that is based on the alternative model proposed in~\cite{becker2019switching}.
This is a model that gives us not only a good dynamics simulation from raw sensor data, but can also be used as a filter for online state estimation.
It is a fully probabilistic model, capturing uncertainty in its state estimation, transition and emission.
It is generally structured as a hierarchical recurrent VAE where one level of latent variable approximates the current state while the other level helps by the determining transition model.
For inference, the authors propose a time-factorised approach with a specific computational structure that reuses the generative model.
We found this structure to be critical for learning a good dynamics model.

\subsection{Generative Model}
The assumed generative model, as depicted in \Cref{fig:generative}, of our data is
\begin{align}
\label{eq:generative-model}
\begin{split}
\p{x_{1:T}}{u_{1:t-1}} &= \int_{z_{1:T}} \int_{s_{2:T}} \p[\xi]{x_t}{z_t} \\
& \p[\xi]{z_t}{z\tm, s_t, u\tm} \: \p[\xi]{s_t}{s\tm, z\tm, u\tm},
\end{split}
\raisetag{32pt}
\end{align}
where $z_t$ and $s_t$ are two levels of latent variables.
The former describes the current state while the latter determines the transition dynamics to the next state $z_{t+1}$.
The generative model consists of the 3 components: likelihood model $\p[\xi]{x_t}{z_t}$, transition $\stran$ and transition $\ztran$.

\subsubsection{Likelihood model}
For our likelihood model, we assume a diagonal multivariate Gaussian distribution whose parameters are predicted by a neural network. 

\subsubsection{Transition over $s_t$}
The latent states $s_t$ determine the transition over latent states $z_t$.
They do not influence the reconstruction of the current observation $x_t$, but only those of future observations indirectly through influencing $z_{>t}$.
We model them as Gaussian distributions and their transition is parametrised by a neural network:
\begin{equation}
\begin{gathered}
\stran = \gauss{\mu_{s_t}, \sigma_{s_t}^2}  \\
\text{where } \; [\mu_{s_t}, \sigma_{s_t}^2] = g_\xi(s\tm, z\tm, u\tm).\\
\end{gathered}
\end{equation}

\subsubsection{Transition over $z_t$}
Latent states $z_t$ represent the current state of the system.
Here, we found that neural network transition is not ideal for various reasons.
Mostly, neural networks are very powerful as they can model very complicated transitions that make it harder to learn a useful latent state $z_t$. 
Instead, we learn locally linear dynamics which have many desirable properties: local smoothness, greater generalisation and smaller data demand.
For learning such dynamics, we maintain a set of $m$ base matrices $\lbrace \left( A^{(i)},B^{(i)},C^{(i)} \right)| \: \forall i,\; 0 < i < m \rbrace$ that are summed depending on variables $s_t$.
Formally, this is defined as 
\begin{equation}
\begin{gathered}
\label{eq:linear-combination}
A_t = \sum_{i=1}^m{\alpha_t^{(i)} A^{(i)}}\quad \text{with} \quad \alpha_t = \sigma(Ws_t + b).
\end{gathered}
\end{equation}
Matrices $B_t$ and $C_t$ are computed analogously to $A_t$.
All entries of the base matrices are initialised at the beginning of training by a scaled normal distribution.
The transition is then defined as
\begin{equation}
\begin{aligned}
\ztran &= \gauss{\mu_{z_t}, \sigma_{z_t}^2} \\
\text{where } \; \mu_{z_t} &= A_\xi(s_t) z\tm + B_\xi(s_t) u\tm\\
\text{and } \; \sigma_{z_t}^2 &= C_\xi(s_t).\\
\end{aligned}
\end{equation}

\subsection{Inference Model}
The inference model is key to learning a good latent representation and latent dynamics, and later for online filtering.
\cite{becker2019switching} achieve this by formulating the inference scheme as a local optimisation around the prior prediction where information from the observation may adjust but not completely overwrite the prior prediction.
This allows the gradient of the reconstruction loss to propagate through the transition dynamics.
They split the inference model into two parts: 1) the transition model $\qtran$ and 2) an inverse measurement model $\qmeas$.
This split allows them to reuse the generative transition model $\ztran$ in place of $\qtran$. 
Then, both components give independent predictions about the new state $z_t$ which are combined in the following manner:
\begin{gather}
\begin{aligned}
\q[\psi]{z_t}{z\tm, s_t, x_t, u_{\tminus 1}} &= \gauss{\mu_q, \sigma_q^2} \\
\text{where } \;\;\gauss{\mu_q, \sigma_q^2} \propto \qmeas &\times \ztran \\
\text{and } \;\qmeas &= \gauss{\mu_{\mathrm{meas}}, \sigma_{\mathrm{meas}}^2} \\
\text{where } \; [\mu_{\mathrm{meas}},\sigma_{\mathrm{meas}}^2] &= h_\psi(x_t).
\end{aligned}
\raisetag{58pt}
\end{gather}
The densities are multiplied resulting in another Gaussian density
\begin{equation}
\begin{aligned}
\mu_q &= \frac{\mu_{\mathrm{trans}} \sigma_{\mathrm{meas}}^2 + \mu_{\mathrm{meas}} \sigma_{\mathrm{trans}}^2}{\sigma_{\mathrm{meas}}^2 + \sigma_{\mathrm{trans}}^2},\\
\sigma_q^2 &= \frac{\sigma_{\mathrm{meas}}^2 \sigma_{\mathrm{trans}}^2}{\sigma_{\mathrm{meas}}^2 + \sigma_{\mathrm{trans}}^2}.
\end{aligned}
\end{equation}
These parameters form the latent state distribution of $z_t$.
This inference structure is depicted in \Cref{fig:fusion} and an identical scheme is used for latents $s_t$.
\\\\
A weakness of this scheme is that it does not look at future observations. 
In particular, finding an initial state has to be decoupled from this filtering procedure and requires a separate parameterisation:
\begin{equation}
\begin{aligned}
\q[\psi]{h}{x\tsub{1}{T},u\tsub{1}{T}} &= \gauss{h;\mu_h,\sigma_h^2} \\
\text{where} \quad [\mu_h,\sigma_h^2] &= i_\psi(x\tsub{1}{T},u\tsub{1}{T})\\
z_1 &= t_\psi(h).
\end{aligned}
\end{equation}
On the robot, we perform this initial inference based on a small sliding window together with a few filtering steps to estimate the current state $z_t$.

\subsection{Objective}
The objective to maximise for this model is the ELBO:
\begin{gather}
\label{eq:loss}
\begin{aligned}
&\loss[\xi,\psi]{x\Ts}{u\Ts} = \sum_{t=1}^T \Big( \expc[\q[\psi]{z_t}{\cdot}]{\log \p[\xi]{x_t}{z_t}} \\ 
&- \expc[\q[\psi]{z_{t-1}, s_{t-1}}{\cdot}]{\kl{\q[\psi]{s_t}{\cdot\:}}{\p[\xi]{s_t}{s\tm,z\tm,u\tm}}}\\
&- \expc[\q[\psi]{z_{t-1}, s_t}{\cdot}]{\kl{\q[\psi]{z_t}{\cdot\:}}{\p[\xi]{z_t}{z\tm,s_t,u\tm}}} \Big).
\end{aligned}
\raisetag{58pt}
\end{gather} 
We learn the parameters of our model end-to-end by backpropagation through time and we  approximate the expectations by a single sample.
Since our latent variables are modelled by diagonal multivariate Gaussian distributions, the KL-divergences can be computed analytically.

\subsection{Learning a Reward Function}
To simulate a reinforcement learning setting, on top of the dynamics simulation, we also require a reward function. 
In practice, the reward function is often based on the entire state or even some additional quantities like contact points~\cite{brockman2016openai}.
However, we would also like to learn in settings with only partial observations where not all of these quantities are available.
Thus, we can not directly compute the rewards based on predicted observations of an imagined trajectory.
Instead, we must build a model of the reward function which is based on the learnt latent state.
Here, even in the partially observed cases, we expect the latent space to contain all necessary information for prediction and thus also for modelling a reward function.
We introduce a separate neural network for reward prediction $\p[\xi]{r_t}{z_t, u_t}$ and extend the general loss of \Cref{eq:loss} by a term for the reward function:
\begin{equation}
\label{eq:reward-regression-loss}
\loss[\textrm{Reward}]{r\Ts}{u\Ts} = \sum_{t=1}^T \expc[z_t \sim q_\psi]{\log{\p[\xi]{r_t}{z_t, u_t}}}.
\end{equation}
Note that we stop the gradient such that this term does not shape the latent space as this changes and possibly simplifies the underlying problem significantly.
Additionally, this would render the model non-transferable to different tasks.

\section{Learning a Controller}
\label{sec:learning-a-controller}
Generally, our algorithm falls into the category of Actor-Critic methods as introduced in section~\ref{sec:background-rl}, meaning we learn both a parametrised policy (actor) and value function (critic) that evaluates the actor.
In terms of design choices, having a generative model of the environment gives us every freedom in shaping our policy optimisation scheme.
However, for policy and value function optimisation, we limit ourselves to data generated by the learnt simulator for optimisation. 
The observed data is thus only used for fitting the generative model, but not directly for learning the controller.
This is to demonstrate what is possible with a purely model-based approach which is clearer to evaluate if we do not make use of real-world rollouts for policy optimisation.
A hybrid approach, where real-world data is used in an off-policy fashion, is also possible and the extension for that is quite straightforward using importance sampling (see \eg \cite{heess2015learning}).

\begin{algorithm}[t]
\caption{Model-based Actor-Critic}
\label{alg:train}
\begin{algorithmic}[1]
\State Initialise model parameters $\theta, \phi, \xi, \psi$ randomly.
\For{each data collection}
\State \texttt{\textbackslash\textbackslash $\:$ Collect data in real environment}
\For{each episode}
\For{t=1..T}
\State Get state estimate $z_t \sim \q[\psi]{z_t}{x_t, z_{t-1}, u_{t-1}}$\footnotemark
\State Evaluate policy $u_t \sim \policy $
\State \texttt{\textbackslash\textbackslash $\:$ Execute control in env.}
\State $x_{t+1}, r_t \leftarrow \texttt{execute}(u_t)$
\EndFor
\State Add episode $(\xseq, \useq, \rseq)$ to data set $\mathcal{D}$.
\EndFor
\State
\State \texttt{\textbackslash\textbackslash $\:$ Optimise model and controller}
\For{each iteration}
\State \texttt{\textbackslash\textbackslash $\:$ Fit dynamics model}
\State Sample $b$ episodes ${(\xseq, \useq, \rseq)} \sim \mathcal{D}$
\State Update model parameters $\xi, \psi$ based on \Cref{eq:loss}
\State \texttt{\textbackslash\textbackslash $\:$ Fit actor and critic}
\State Imagine a batch of trajectories starting from a random subset of filtered states
\State Update policy $\theta$ with $\nabla_\theta$ of \Cref{eq:n-step-policy-objective}
\State Update value function $\phi$ with $\nabla_\phi$ of \Cref{eq:value-function-objective}
\State Update target value func. $\phi' \gets \alpha_{\phi'} \phi + (1 - \alpha_{\phi'}) \phi'$
\EndFor
\EndFor
\end{algorithmic}
\end{algorithm}

Going more into detail, \Cref{alg:train} sketches a high-level overview of the algorithm.
After collecting an initial data set and optimising a preliminary dynamics model, we start with the main optimisation loop in line 15 and onwards.
For optimising the controller we root every dreamed rollout, \ie a rollout within the latent dynamics model, in a real-world experience by randomly choosing points from the data set and computing the corresponding latent state by filtering up to that point.
From then on we follow the policy for a fixed and small number of steps in our latent dynamics model.
This rollout is used to optimise both actor and critic via stochastic analytic gradients.

The length of the rollout is a trade-off between exploiting the model's internal structure for gradient computation and limiting the impact of the model's accumulating prediction error.
Long imagined trajectories are generally not necessary for two reasons. 
First, we use a learnt value function as a critic to evaluate the policy's performance instead of relying on a Monte Carlo estimate of the reward.
Second, we are free to start a rollout at any observed state that we have ever seen before, which is in contrast to \eg a real-world setup where the robot always resets to the same starting position.
On the other hand, imagining just a single step limits our use of the model severely.
Generating data this way has numerous advantages: we are always on-policy, we can create arbitrarily many experiences and have no need for a replay buffer.

Note that the various updates of model, policy and value function are not tied to each other and may be performed at independent intervals.
Others \cite{fujimoto2018addressing} have noted that updating the actor less frequently than the critic may be beneficial.

\begin{figure*}[t]
  \centering
  \subfloat{\includegraphics[width=\columnwidth]{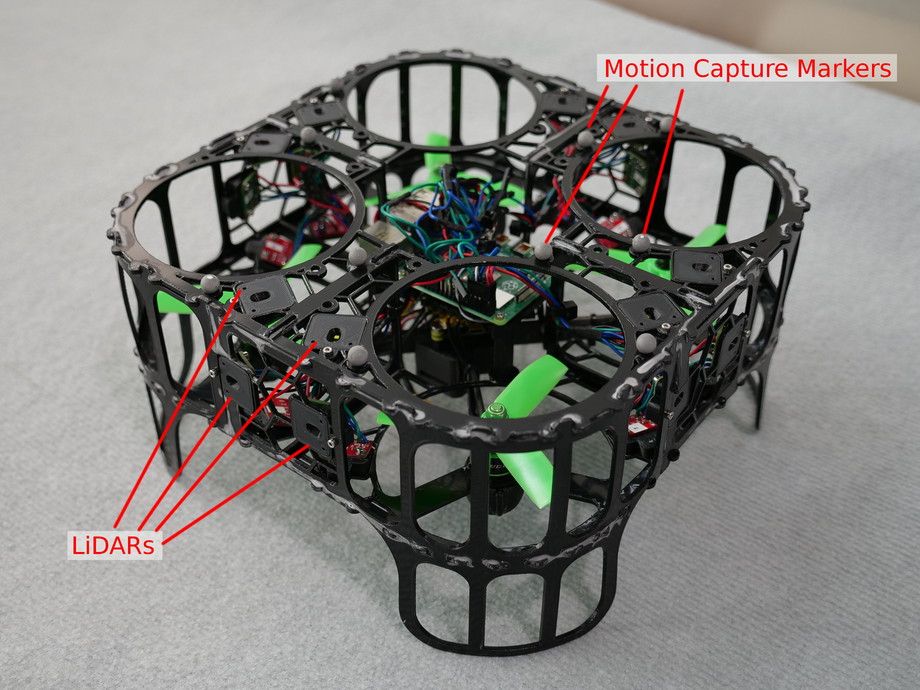}}%
  \hfill
  \subfloat{\includegraphics[width=\columnwidth]{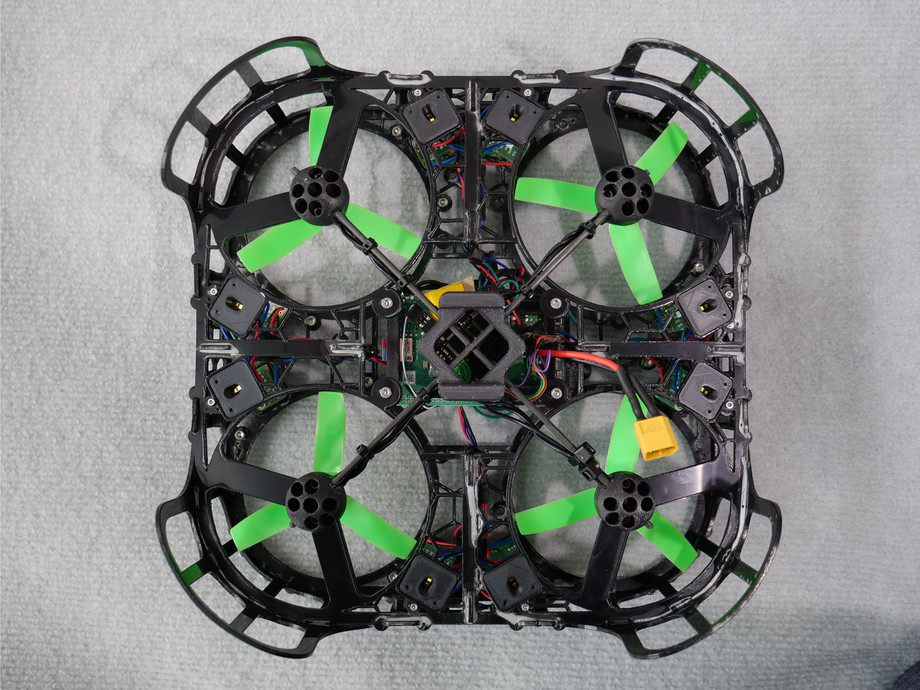}}%
  \caption{Our quadrotor with 24 LiDARs (8 of them facing to the top, side and bottom), motion capture markers and a Raspberry Pi 4. The flight controller including the IMU is placed directly beneath the Raspberry Pi. The frame makes the drone withstand collisions with the walls or floor.}
  \label{fig:drone}
\end{figure*}

\subsection{Value Estimation}
\label{ssub:value_function}
As mentioned, we optimise the value function based on on-policy rollouts in our latent dynamics model, where a rollout is formally defined as 
\begin{equation}
\label{eq:trajectory}
\tau_{\theta, \xi} \sim \p(z_1) \prod_{t=1}^{H-1} \policy \apxtransition.
\end{equation}
Note that for notational convenience we subsume our hierarchical latent space consisting of $\zseq$ and $\sseq$ into a single state sequence $\zseq$ for the remainder of this paper.
Given these rollouts, state values can be estimated in numerous ways, most commonly by minimising the \emph{temporal difference} error (TD-error)~\cite{sutton1998introduction}:
\begin{equation}
\label{eq:td-error}
\expc[\tau_{\theta, \xi}]{\left( \apxvalue - \left( \apxreward  + \gamma \apxvaluenext \right) \right)^2}.
\end{equation}
For faster convergence, we make use of an n-step variant~\cite{watkins1989learning}, which  reduces the bias at the cost of increased variance:
\begin{equation}
\begin{aligned}
\label{eq:value-function-objective}
&\expc[\tau_{\theta, \xi}]{\left( \apxvalue - y_{t} \right)^2}\\
\text{with} \quad  &y_{t} = \sum_{i=0}^{H-1} \gamma^{i} \r[\xi]{z_{t+i}, u_{t+i}}  + \gamma^{H} \apxvaluenextNtarget. \\
\end{aligned}
\end{equation}
Since we sample from both policy and latent dynamics model, we can minimise this objective \wrt parameters $\phi$ by computing analytic stochastic gradients using the reparameterisation trick.
To stabilise this regression, we introduce a \textit{target} value network $\apxvaluenextNtarget$ parametrised by $\phi'$~\cite{mnih2015human}.
The target network's parameters $\phi'$ are slowly updated towards $\phi$ using a small learning rate $\alpha_{\phi'} \ll 1$ and thus the target values of the regression are changing more slowly:
\begin{equation}
\phi' \gets \alpha_{\phi'} \phi + (1 - \alpha_{\phi'}) \phi'.
\end{equation}
We parameterise the value function by a neural network.
\footnotetext{Note that for notational convenience we subsume our hierarchical latent space consisting of $\zseq$ and $\sseq$ into a single state sequence $\zseq$ for the remainder of this paper.}

\begin{table}[t]
\renewcommand{\arraystretch}{1.25}
\caption{Drone Hardware Components}
\begin{center}
\begin{tabular}{ll}
\hline
Component & Description \\
\hline
\multicolumn{2}{l}{\textbf{Flight Electronics}} \\
Flight Controller & CLRACING F7 \\
Motors & $4 \times$1808-2600kv \\
ESC & 4-In-1 30A with BLHeli-S \\
Propellers & $4 \times 4"$ Bull Nose, $3$ Blades \\
Battery & 4S LiPo, 1400mAh, 65C \\
\multicolumn{2}{l}{\textbf{Perception \& Computation}} \\
Onboard & Raspberry Pi 4 \\
LiDAR & $24 \times$VL53L1X \\
IMU & ICM-20602 \\
\hline
\end{tabular}
\label{tab:drone-hw}
\end{center}
\end{table}

\subsection{Policy}
Using the critic, we can optimise our policy, which we choose to be represented as a diagonal multivariate Gaussian distribution, \ie
\begin{equation}
\policy \sim \gauss{\mu_\theta(z_t),\sigma_\theta(z_t)}.
\end{equation}
In general, any distribution where the reparameterisation trick is applicable may be used instead.
Similar to optimising the value function, we make use of short policy rollouts within the learnt dynamics model.
These rollouts do not need to be the same or be of the same length as used for learning the critic.
As the general objective for policy optimisation in Actor-Critic methods, one chooses the gradient of the critic, \ie
\begin{equation}
\label{eq:policy-objective}
\nabla_\theta \expc[\tau_{\theta, \xi}]{\apxreward  + \gamma \apxvaluenext}.
\end{equation}
Here again, we make a slightly different compromise between Monte Carlo estimation and relying on the critic by choosing to sum up observed rewards before truncating the series with the (discounted) critic of the trajectory's terminal state:
\begin{equation}
\label{eq:n-step-policy-objective}
\expc[\tau_{\theta, \xi}]{\sum_{i=0}^{H-1} \gamma^{i} \r[\xi]{u_{t+i},z_{t+i}}  + \gamma^{H} \apxvaluenextN}.
\end{equation}
A typical rollout length used in our experiments is between $3$ and $10$.
Analogously to the optimisation of the critic, this objective can be maximised \wrt policy parameters $\theta$ by backpropagation using the reparameterisation trick.

\section{Experimental Platform: Our Drone}
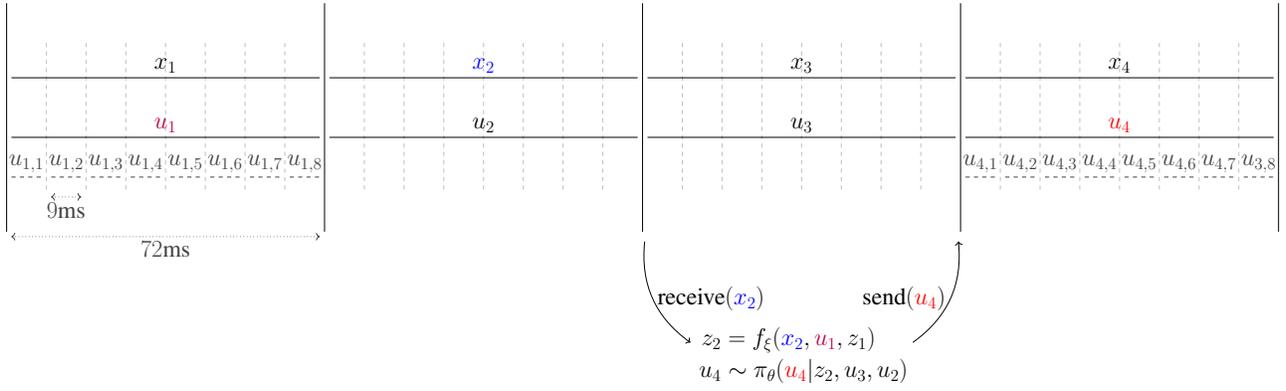
\begin{figure*}[t]
  \centering 
  \resizebox {.95\textwidth} {!} {
    \begin{tikzpicture}
\tikzstyle{every node}=[font=\LARGE]
\def\modelintervals{2}
\def\lastinterval{4}
\def\modelstepwidth{8}
\def\modelstepheight{6}

\foreach \step in {0,...,4}{
	\node at (\step * \modelstepwidth, \modelstepheight) (t\step) {};
	\node at (\step * \modelstepwidth, 0) (b\step) {};
	\node at (\step * \modelstepwidth, \modelstepheight - 2) (x\step) {};
	\node at (\step * \modelstepwidth, \modelstepheight - 3.5) (u\step) {};
	\draw (t\step) edge (b\step);

	\ifthenelse{\step < 4}{
		\foreach \substep in {0,...,7}{
			\node at (\step * \modelstepwidth + \substep, \modelstepheight - 1) (t\step\substep) {};
			\node at (\step * \modelstepwidth + \substep, 1) (b\step\substep) {};
			\node at (\step * \modelstepwidth + \substep, \modelstepheight - 4.5) (u\step\substep) {};
			\ifthenelse{\substep > 0}{
				\draw (t\step\substep) edge[dashed,lightgray] (b\step\substep);
			}{}
		}
	}{
		\node at (4 * \modelstepwidth, \modelstepheight - 4.5) (u40) {};
	}
	
}

\foreach \substep in {1,...,7}{
	\pgfmathsetmacro\prev{\substep - 1}
	\draw[darkgray,dashed] (u0\prev) -- (u0\substep) node [midway,above]{$u_{1,\substep}$};
}
	\draw[darkgray,dashed] (u07) -- (u10) node [midway,above]{$u_{1,8}$};
\foreach \substep in {1,...,7}{
	\pgfmathsetmacro\prev{\substep - 1}
	\draw[darkgray,dashed] (u3\prev) -- (u3\substep) node [midway,above]{$u_{4,\substep}$};
}
\draw[darkgray,dashed] (u37) -- (u40) node [midway,above]{$u_{3,8}$};

\draw[darkgray,<->,dotted] (b0) -- (b1) node [midway,below]{$72\text{ms}$};
\draw[darkgray,<->,dotted] (b01) -- (b02) node [midway,below]{$9\text{ms}$};

\draw (x0) -- (x1) node [midway,above]{$x_1$};
\draw (u0) -- (u1) node [midway,above, purple]{$u_1$};
\draw (x1) -- (x2) node [midway,above, blue]{$x_2$};
\draw (u1) -- (u2) node [midway,above]{$u_2$};
\draw (x2) -- (x3) node [midway,above]{$x_3$};
\draw (u2) -- (u3) node [midway,above]{$u_3$};
\draw (x3) -- (x4) node [midway,above]{$x_4$};
\draw (u3) -- (u4) node [midway,above,red]{$u_4$};

\node at (20, -3) (agent) {$\begin{aligned}
z_2 &= f_\xi(\textcolor{blue}{x_2},\textcolor{purple}{u_1},z_1) \\
u_4 &\sim \pi_\theta(\textcolor{red}{u_4}|z_2,u_3,u_2)
\end{aligned}$};

\path[every node/.style={font=\sffamily\small}]
	(b2) edge[->, bend right] node [right,font=\LARGE] {$\text{receive}(\textcolor{blue}{x_2})\;$} (agent)
	(agent) edge[->, bend right] node [left,font=\LARGE] {$\text{send}(\textcolor{red}{u_4})$} (b3);

\end{tikzpicture}
  }%
  \caption{Timing of the control loop. The big intervals denote high-level control where we update our state estimation and compute a new control input based on our learnt policy. The smaller $9$ms intervals represent the communication interval with the flight controller. Here, we send a control signal and receive filtered IMU data.}
  \label{fig:timing}
\end{figure*}

We built our own quadrotor (see \Cref{fig:drone}) from scratch as learning autonomous flight starting from random exploration places unique demands on the hardware. 
In particular, we fitted our quadrotor with a robust polyamide frame, allowing it to bump into walls or the ground while staying fully operational.
This simplifies initial exploration and deployment of (preliminary) policies.
Moreover, this frame also allows for mounting various sensors such as cameras or LiDARs.
For this work, the drone is equipped with $24$ VL53L1X time-of-flight sensors which are connected via I2C directly to a Raspberry Pi 4 which performs all computing tasks required for our entire model.
In practice, the LiDARs give us readings between 0.05--3.50m and take up to $50$ms to measure.
The drone is further equipped with a CLRACING F7 flight controller which comes with an ICM-20602 inertial measurement unit (IMU) that combines a three-axis gyroscope and a three-axis accelerometer.
Similar to the LiDARs, the flight controller is connected to the Raspberry Pi.
Lastly, the drone frame is fitted with motion capture markers for an external tracking system.
Overall the drone weighs $640$g without the battery, which weighs another $181$g.
An overview of all key hardware components is given in \Cref{tab:drone-hw}.

\subsection{Environment}
\begin{figure}[t]
  \centering 
  \includegraphics[width=\columnwidth]{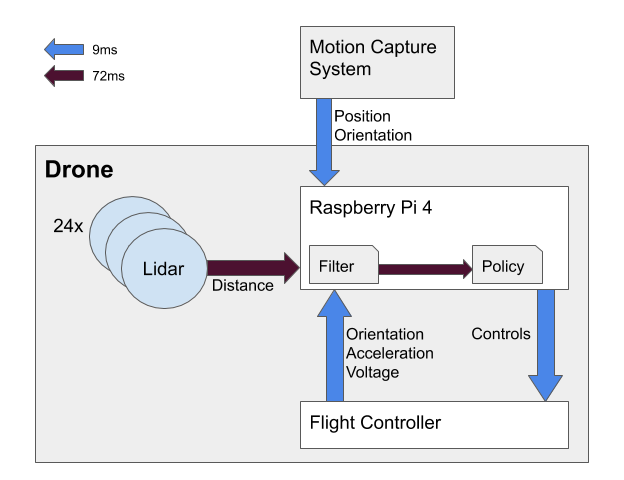}%
  \caption{Visualisation of communication between sensors, onboard computational hardware and external motion capture system. Sending of motion capture data is done over WiFi using Open Robot Communication~\cite{frank2019orc}.}
  \label{fig:drone-component-communication}
\end{figure}

The drone was placed in a $3.0\text{m} \times 3.0\text{m} \times 2.4\text{m}$ cage with acrylic glass walls (see \Cref{fig:drone-trajectory}).
We artificially limited the drone's throttle such that it cannot fly higher than the visible foil at the walls which end at $1.2$m.
The foil's main purpose is to improve the LiDAR readings.
The cage is fitted with a motion capture system (OptiTrack) which operates at up to $120$Hz.
Its data is relayed to the drone over WiFi using  Open Robot Communication (ORC)~\cite{frank2019orc}.
An overview of the various components and their communication is visualised in \Cref{fig:drone-component-communication}.

\subsection{High and Low-Level Controller}
Our control scheme consists of a high and low-level controller, where only the high-level controller is learnt (as described in \Cref{sec:learning-a-controller}). 
The high-level controller is a thrust-attitude controller, \ie it defines throttle, roll, yaw and pitch commands which are then translated to motor torques.
For low-level control we use angle mode of Betaflight 3.5.6~\cite{betaflight} on our flight controller which keeps the drone stable when no control inputs are sent.
In this mode, roll and pitch inputs are translated into fixed desired angle of the drone and a throttle application of 0 approximately keeps the drone hovering.
In practice, this is not quite the case due to variation of individual battery performance and battery charge.
When no (\ie 0) attitude commands are sent to the flight controller, it will level itself.
This is mainly useful for initial exploration as it simplifies the policy design greatly, but is admittedly also helpful for hovering at a goal position after it has been reached.
This general setting is identical to how humans typically learn to operate a drone at the beginning.

\begin{table}[t]
\renewcommand{\arraystretch}{1.25}
\caption{Drone Sensor Streams}
\begin{center}
\begin{tabular}{lccc}
\hline 
Sensor Stream & Dimensions & \specialcell{Measurement\\Period [ms]} & \specialcell{Effective\\Dimensions}  \\
\hline 
LiDAR & 24 & 72 & 24\\
IMU orientation & 3 & 9 & 24\\
IMU acceleration & 3 & 9 & 24 \\
Battery Voltage & 1 & 72 & 1 \\
(Simulated) Compass & 1 & 9 & 8 \\
Mocap Drone Position & 7 & 9 & 56 \\
Mocap Drone Velocity & 3 & 9 & 24 \\
Mocap Goal Marker & 7 & 72 & 7 \\
\hline 
\end{tabular}
\label{tab:drone-sensor-streams}
\end{center}
\end{table}

Since our learnt high-level controller sends thrust-attitude commands to the flight controller, there exists an underlying PID controller that translates them into motor currents.
This PID controller being limited in its operational speed, we would like to avoid a jittery control signal---something that is not enforced by our Gaussian policy.
We therefore introduce a small moving average over the last 6 values (or 54ms) to smoothen out the control signal.
This also ensures that our policy stays within the data distribution of our initial exploration policy which is important for valid model predictions.
Further, the control signal has to obey certain control limits of the real system.
We enforced this by transforming the control input by a tanh function and then scaling it to the required interval.

The low-level controller runs at $2$kHz while the high-level controller sends control commands every $9$ms.
Since computation of our neural state estimation and policy takes longer than $9$ms, we run the high-level controller at $72$ms intervals which predicts the next $8$ actions in an open-loop fashion.
This control loop is shown in \Cref{fig:timing}.

Finally, when applying RL methods, it is often wrongfully assumed that the agent's state does not change during action selection~\cite{ramstedt2019real,travnik2018reactive,walsh2009learning}.
In practice, this is often not disastrous as the time required for action selection is negligible.
We follow~\cite{ramstedt2019real} which allows an agent exactly one time-step to select an action and where the original MDP is augmented by the previous action(s).
Thus, the policy actually takes the form
\begin{equation}
\policydist[\theta]{u_{t+1}}{z_t, u_{t}, u_{t-1}}.
\end{equation}
This recovers the theoretical applicability of the underlying mathematical framework.

\section{Experiments}
We showcase our methodology in various variations of the same scenario: our quadrotor flying from any position to a randomly placed goal marker in an enclosed environment.
This scenario was completed in two settings with either disabled or enabled yaw actuation.
In both settings, drones using different subsets of the available sensory information were compared to each other.

\subsection{Setup}
\label{ssub:experimental_setup}

\subsubsection{Drone Configurations}

\begin{figure*}[t]
  \centering
  \subfloat[]{\includegraphics[width=\columnwidth]{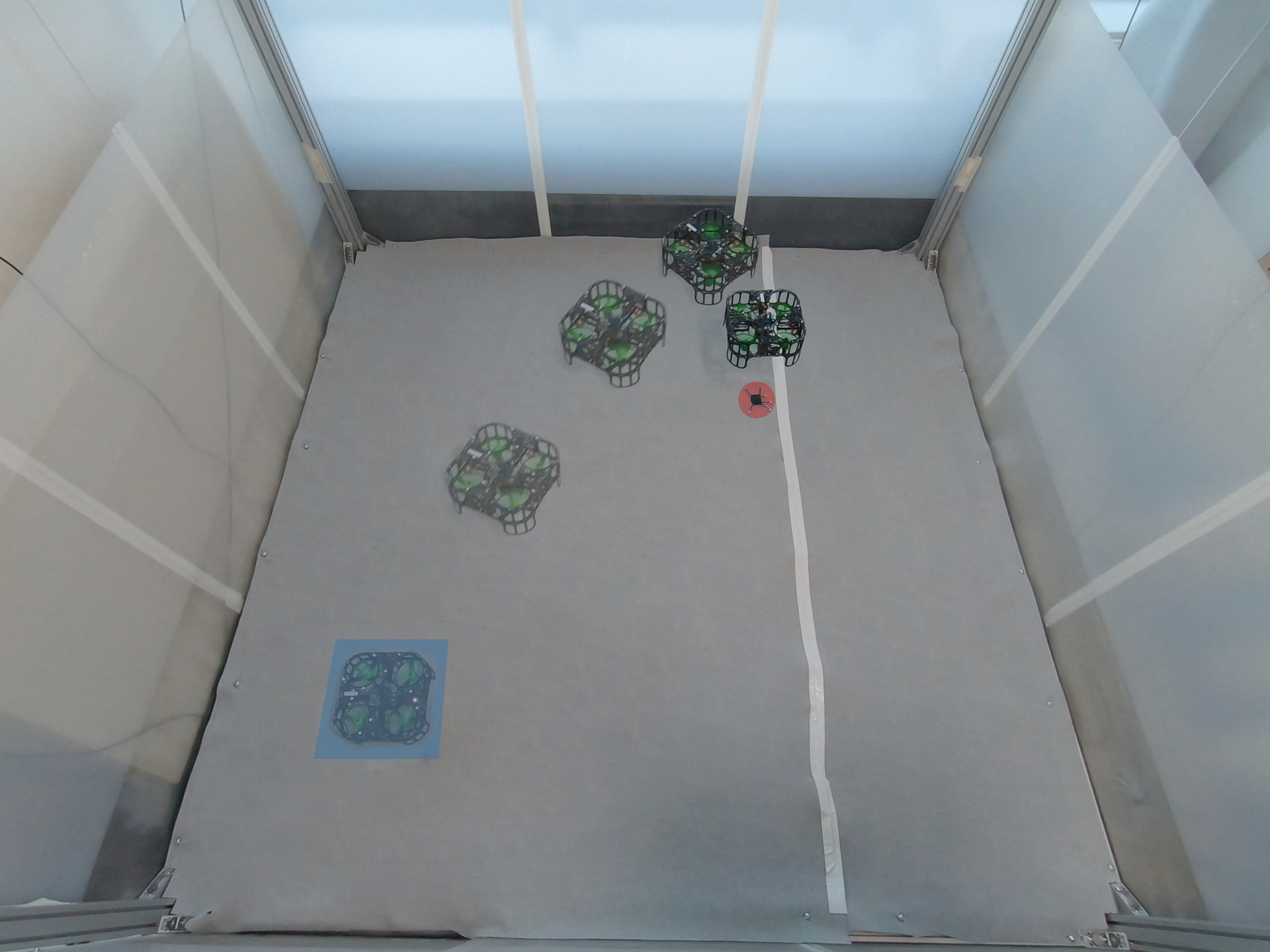}
  \label{fig:drone-trajectory-real}}%
  \hfill
  \subfloat[]{\includegraphics[width=0.82\columnwidth]{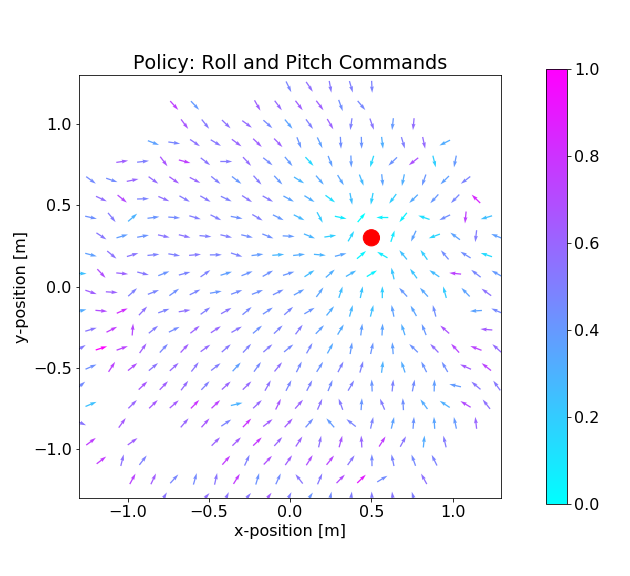}
  \label{fig:quiver-policy}}
  \hfill\null
  \caption{(a) Shows the drone lifting off and flying to the marked goal in the upper right section of the cage. (b) Shows a corresponding quiver plot looking at the cage from a birds perspective and indicates the policy's roll and pitch commands averaged over a validation data set given the goal position of (a). The arrow's colour denotes the relative magnitude of the control input. Video: \url{https://youtu.be/e5buJL_DYgA}.}
  \label{fig:drone-trajectory}
\end{figure*}

We compared three different drone configurations, using only part of all sensory information as listed in \Cref{tab:drone-sensor-streams}. 
Note that some sensors give measurements more frequently than others (see also \Cref{fig:drone-component-communication}), but our model works at a fixed interval length of $72$ms.
If multiple measurements are available, we simply provide our model with all of them.
Our three configurations are defined as follows:
\begin{itemize}
\item{\textit{MocapVel} observes motion capture estimates for position and Cartesian velocity. 
Velocity is computed using backward differences on filtered motion capture positions.}
\item{\textit{Mocap} observes only motion capture positions, but no velocity.
Thus, it only observes a partial state and has to infer its velocity indirectly.}
\item{\textit{LiDARVel} uses only onboard sensors, including all 24 LiDARs and the IMU, but no motion capture position or velocity. 
However, to make unique and global position identification at all possible, we simulate a compass using our motion capture system, \ie we provide the drone's z-orientation in the global motion capture frame.
Last, we supply the model with a LiDAR-based Cartesian velocity estimate based on averaging estimates of LiDARs pointing in the same or opposite direction.}
\end{itemize}
Further, all drones receive a current battery voltage measurement which is updated approximately every $400$ms and the motion capture position of the goal marker as an observation.

\subsubsection{Data Collection}
Exploration in this setting is a difficult endeavour as we require prolonged actions in one direction for the drone to do something meaningful.
Random control commands, as is often done in RL research, would thus lead to the drone mostly sitting on the floor or being stuck at one of the walls.
This natural behaviour of a random agent in a bounded environment is aggravated by the suction effect near the walls.
On the other hand, it is very easy to accelerate the drone to high speeds by excessively tilting in one direction which leads to uncontrollable flight.
As both scenarios are undesirable, we start out by collecting an initial data set with a PID controller and continue with the learnt policy in subsequent recording steps.

For the initial data set, the PID controller flies to randomly chosen targets.
To facilitate learning, we update each component of the goal, which consists of $(x,y,z)$-position and $z$-orientation, independently after a randomly drawn interval length (on average around $1$s).
This leads to more decorrelated actuation of the control dimensions when compared to changing the entire goal at once.
Note that it is not important for the PID controller to actually reach a target or do a good job at controlling the drone at all.
We merely require varied and stable drone flight for our initial data set to learn a first dynamics model.
After recording an initial data set, we can record subsequent data using the learnt policy.
We collected about $10$ of overall $30$ minutes of flight data using this initial exploration scheme.

\begin{figure*}[t]
  \centering
  \subfloat{\includegraphics[width=0.33\textwidth]{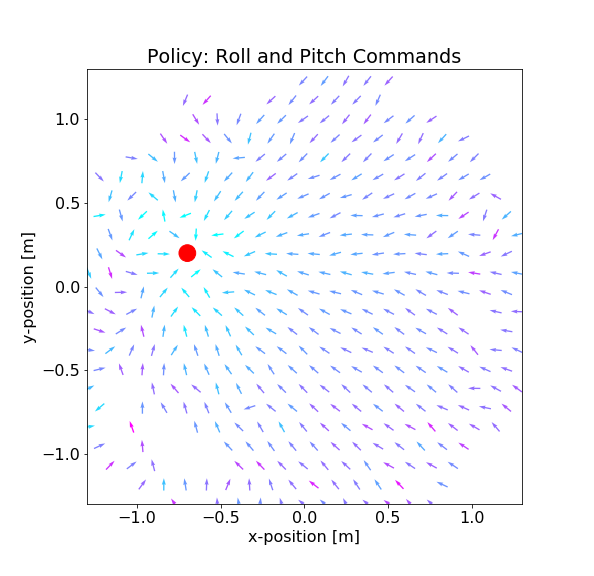}}%
  \hfill
  \subfloat{\includegraphics[width=0.33\textwidth]{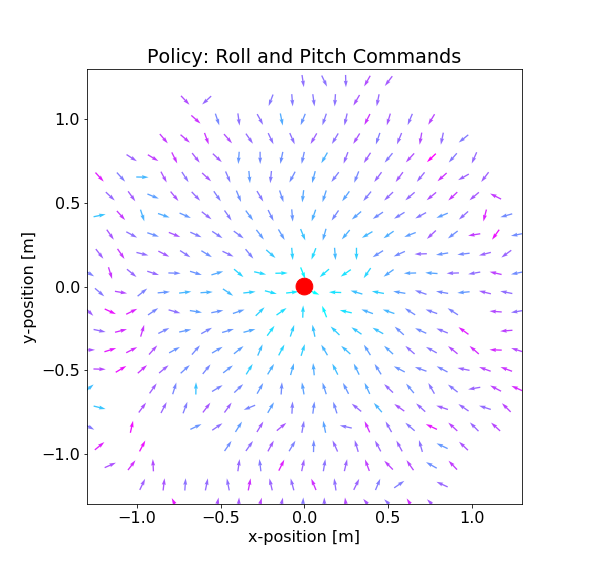}}%
  \hfill
  \subfloat{\includegraphics[width=0.33\textwidth]{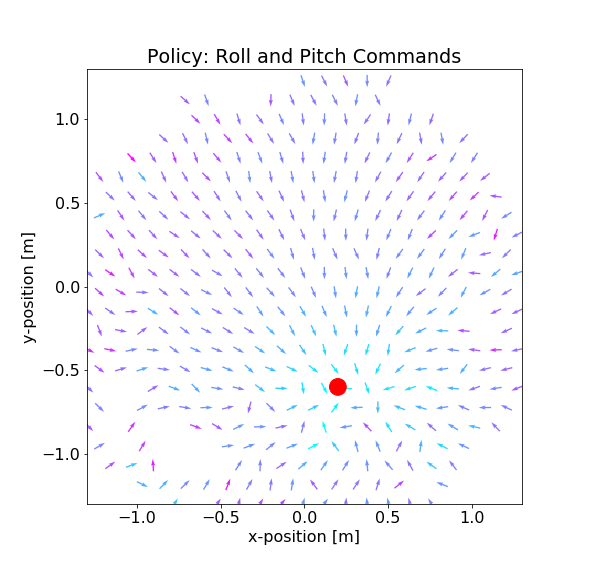}}%
  \caption{These plots showcase a single policy's roll and pitch commands from the birds perspective when conditioned on various goals as marked by the red dot. Data points are taken from a validation set and the policy's actions are averaged over variations in velocity, height and orientation. Some areas of the plots are empty, indicating a lack of data in the validation set.}
  \label{fig:multiple-quiver-plots}
\end{figure*}
\begin{figure*}[t]
  \centering
  \includegraphics[width=\textwidth]{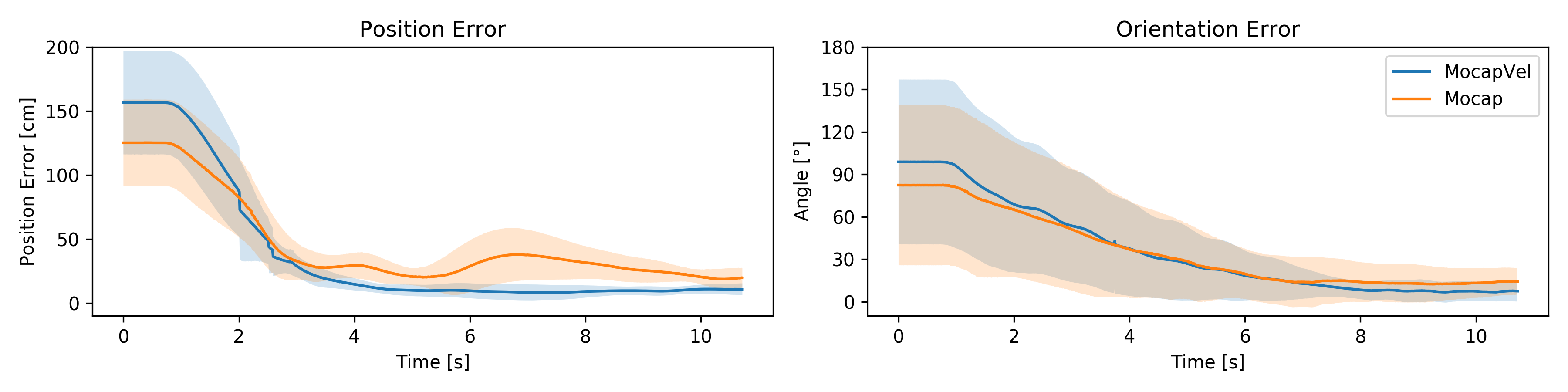}%
  \caption{Comparison of the position and z-orientation error over time averaged across all successful test flights. Shaded area marks one standard deviation. One can see how \textit{Mocap} requires a bit longer to settle, oscillating a bit before converging to a slightly larger error than the fully observable \textit{MocapVel} configuration.}
  \label{fig:yaw-error-over-time}
\end{figure*}

\subsubsection{Data Processing}
Minimising engineering effort being a primary motivation of our work, we wanted to limit data processing requirements as much as possible; nevertheless a few processing steps are necessary.
As is common when working with neural networks, we normalised all input data to the interval $[-1;1]$ based on known sensor limits.

Both the motion capture data and IMU signal are already filtered by their respective sources and we applied no further filtering.
The tracking system gives us the orientation of the drone and goal marker in quaternions.
Quaternions are difficult for neural networks to learn as there are discontinuities and ambiguity in representation.
Instead of learning on them directly, we followed~\cite{zhou2019continuity} and learnt on a unique and smooth 6D representation which is based on 2 columns of the corresponding rotation matrix.
For the motion capture data, we mapped the drone's position from the tracking frame to the drone's frame.
This helps the model in so far as the actions and motion capture data are represented in the same frame of reference.
The goal marker position was not translated or otherwise processed.

LiDAR data is filtered using a weighted moving average over the last 5 measurements, both for computational simplicity and minimal added delay.
Based on this filtered signal, we estimate their rate of change using backward differences over the same window. 
Then, LiDARs pointing in the same or opposite direction are averaged to compute a Cartesian velocity estimate as we found the use of singular velocity estimates to be too noisy for the model.

For training of the sequential latent variable model, we applied a sliding window on the recorded trajectories, each of which is approximately a minute long, that yields subsequences of ca. $3-6$s duration (\ie $40-80$ discrete time steps).
This increases the diversity of observed initial states and accelerates training.
For each subsequence, we augmented the data set by overwriting the recorded goal marker position by a randomly drawn one, increasing the diversity of observed goals.
Accordingly, we recomputed the observed reward to match the new goal.

\subsection{Fly to Marker}
Our goal scenario is for a drone to fly to a marker, \ie from a standstill, it is supposed to lift off, fly and hover at a desired height over a goal marker that may be put anywhere in the cage. 
We present results on two variations of this scenario, one simplified scenario where the drone's yaw actuation was disabled and one where it was not.
This is because some drone configurations were unable to complete the harder scenario successfully.

The global position of the goal marker is observed by the external motion capture system and relayed to the drone.
That goal may be treated just as any other observation and be concatenated to the model input. Alternatively, as we do in our experiments, it may be used directly as a condition for the policy and during training, the reward and value function.
A scenario is completed successfully if the drone hovers within a certain distance and orientation (in the scenario where yaw actuation is enabled) for a duration of $3$s.
Specifically, its distance may not exceed $30$cm and its orientation may not exceed $15$ degrees within the time frame.
If an episode is successful, we compute the average final error over the following second.
These metrics are then used to showcase both the absolute achieved performance and to compare between the different drone configurations.

As a reward function, we define a common one across all scenarios as 
\begin{equation}
\label{eq:reward-definition}
-(p - p_{\textrm{goal}})^2 - 0.1 (\theta_z - \theta_{\textrm{goal}})^2 - 0.1 v^2 - 0.001 u^2,
\end{equation}
where $p[\textrm{m}]$ is the drone's Cartesian position in the global tracking frame, $\theta_z[\textrm{rad}]$ its z-orientation, $v[\textrm{m/s}]$ its Cartesian velocity and $u$ the normalised control input.
The quantities $p_{\textrm{goal}}$ and $\theta_{\textrm{goal}}$ refer to the position and orientation of the goal marker.
This last term is omitted for the scenario with disabled yaw actuation.
In the other scenario, the goal orientation is fixed and not dependent on the goal marker's orientation.
If successful, the white stripe of the drone should always face upwards in the end (see \Cref{fig:drone-trajectory}).

\begin{table}[t]
\renewcommand{\arraystretch}{1.3}
\caption{Fly to Marker: Quantitative Results}
\begin{center}
\begin{tabular}{l|cc|cc}
\multicolumn{5}{c}{\textbf{Yaw Active}} \\
\hline 
\multicolumn{1}{c}{} & \multicolumn{2}{c}{Success} & \multicolumn{2}{c}{Average Final Error} \\
Configuration & Rate & Time [s] & Position [cm] & Orientation [\si{\degree}] \\ 
\hline 
MocapVel 	& $100\%$ & $6.1 \pm 2.81$ & $4.41 \pm 1.74$ & $6.88 \pm 2.80$\\
Mocap 	    & $67\%$ & $10.5 \pm 4.59$ & $5.15 \pm 2.41$ & $7.75 \pm 3.73$\\
\multicolumn{1}{c}{} & \rule{0pt}{3ex} \\
\multicolumn{5}{c}{\textbf{Disabled Yaw / Fixed z-Orientation}} \\
\hline 
\multicolumn{1}{c}{} & \multicolumn{2}{c}{Success} & \multicolumn{2}{c}{Average Final Error} \\
Configuration & Rate & Time [s] & Position [cm] & Orientation [\si{\degree}] \\ 
\hline 
MocapVel 	& $100\%$ & $2.5 \pm 0.32$ & $6.17 \pm 2.48$ & $-$\\
Mocap 		& $100\%$ & $7.6 \pm 3.22$ & $6.53 \pm 3.30$ & $-$\\
LiDARVel 	& $44\%$ & $17.2 \pm 6.68$ & $7.29 \pm 1.52$ & $-$\\
\end{tabular}
\label{tab:results}
\end{center}
\end{table}

\subsubsection{Disabled Yaw}
In this simplified scenario, the drone's yaw actuation was disabled and it was placed axis-aligned in the environment, \ie facing directly in one of the four cardinal directions.
An example of this scenario is shown in \Cref{fig:drone-trajectory-no-yaw}.
We report our main metrics in \Cref{tab:results} which were averaged over 9 different start and goal positions. 
The same start and goal positions were used for each drone configuration.

\begin{figure}[t]
  \centering
  \includegraphics[width=\columnwidth]{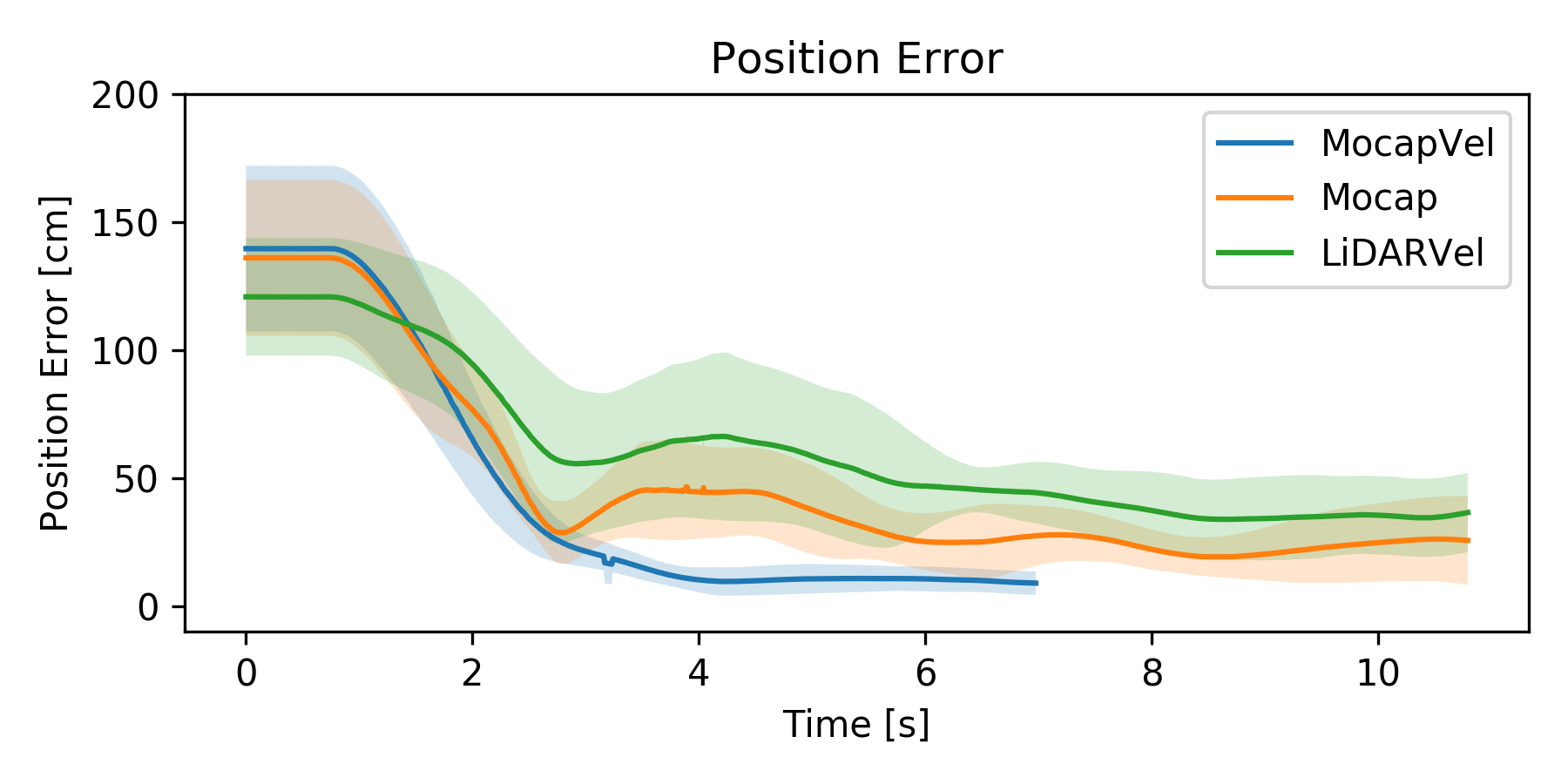}%
  \caption{Comparison of position error over time in the scenario where yaw actuation is disabled.}
  \label{fig:noyaw-error-over-time}
\end{figure}

We see that \textit{MocapVel} is successful in all cases, achieving success much faster and with lower error then the other two configurations.
Without an observed velocity (\textit{Mocap}), the learnt controller is still good enough to complete the task, albeit it requires significantly more time and concludes with a slightly higher final error.
Having no velocity observation, it has a much harder time to find a perfect stand still above the goal and oscillates around the goal.
This behaviour can readily be seen in \Cref{fig:noyaw-error-over-time}, displaying the remaining error over time of successful trajectories.

Lastly, while \textit{LiDARVel} possesses a velocity estimate, both its position and velocity measurements are much more noisy than the motion capture data.
Thus, it represents the hardest setting as can be inferred by its lower rate of success.
Here, we found that our learnt controller in all our attempts only flew well while oriented in some of the four possible fixed orientations.
The underlying issue we suspect to be related to the translation from its local position measurement to the global goal frame, but we were unable to resolve it.
Other failures take the form of oscillating around the goal, leaving the desired goal region periodically within $3$s. 
In general, it took much longer until the drone stayed within the defined goal region for a sufficient amount of time and even drifted away from it after achieving a momentary standstill.
We also tried a fourth configuration \textit{LiDAR} where we removed the velocity estimate from the observation and tried to control based on LiDAR-sensors alone, but were unable to complete the scenario.

\subsubsection{Activated Yaw}
When activating yaw actuation, we were only able to get results in the \textit{MocapVel} and \textit{Mocap} configuration.
An example of a successful completion of this scenario is shown in \Cref{fig:drone-trajectory}, where the drone now also has to rotate to specific z-orientation.
Analogously to before, we report final errors in \Cref{tab:results} and show the remaining error of position and orientation over time in \Cref{fig:yaw-error-over-time}.
Here, we can see that \textit{MocapVel} is still successful in all cases, while \textit{Mocap} is only successful in two thirds of the scenarios.
The failures were mainly due to the drone not rotating all the way to the required goal orientation, and rarely due to excessive oscillation around the goal position.
Again, we can see that \textit{Mocap} takes a bit longer to settle at the goal, with slightly higher final errors in both position and orientation.

\begin{figure*}[t!]
\centering
  \subfloat[]{\includegraphics[width=\textwidth]{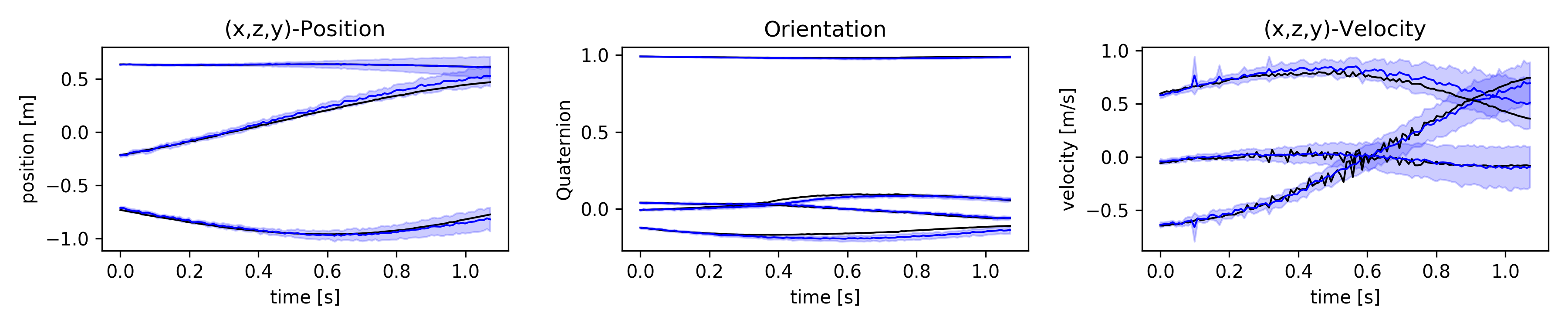}}%
  \hfill
  \subfloat[]{\includegraphics[width=\columnwidth]{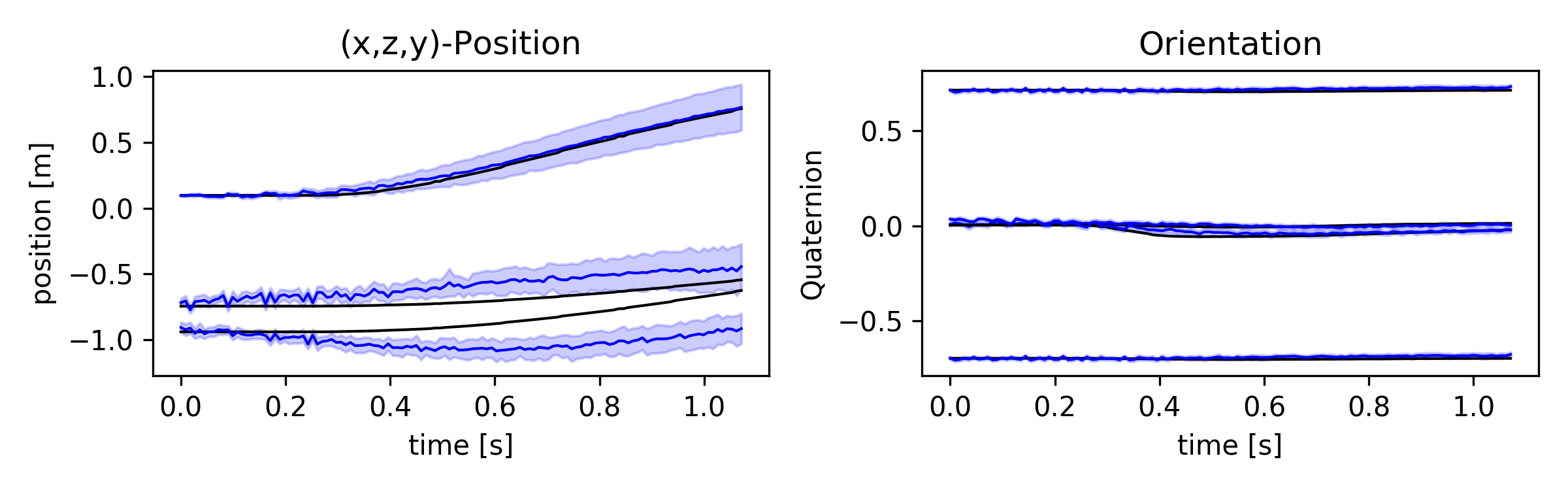}}%
  \hfill
  \subfloat[]{\includegraphics[width=\columnwidth]{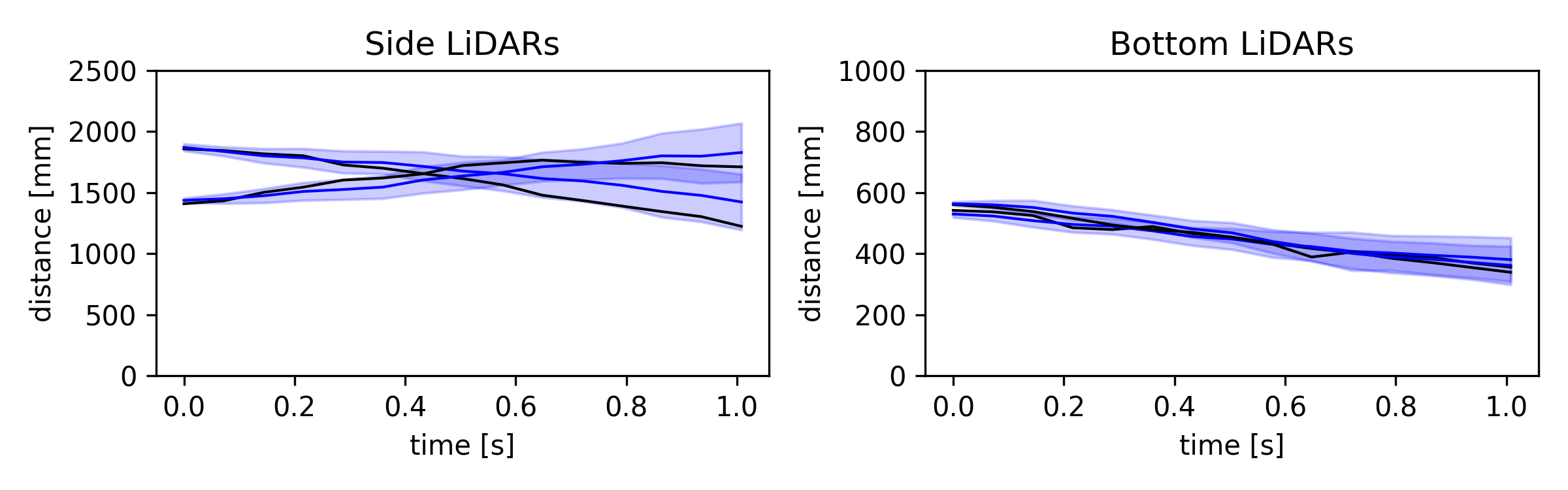}}%
\caption{These plots showcase the predictive performance of our probabilistic model. Conditioned on the a filtered state and future control inputs, they show predictions about the evolution of various sensors over the next second. The black line denotes the data, the blue line denotes the prediction average and the bluely shaded area marks one standard deviation based on 20 predictions. The predictions are based on different drone configurations: (a) \textit{MocapVel}, (b) \textit{Mocap} and (c) \textit{LiDAR}.}
\label{fig:model-quality}
\end{figure*}

We showcase a qualitative evaluation of a single learnt policy in \Cref{fig:multiple-quiver-plots} and how it changes depending on the observed goal.
To get a better impression of the actual performance, we recommend taking a look at our video that showcases both successful and failed episodes of all scenarios and drone configurations: \url{https://youtu.be/e5buJL_DYgA}.\\
Lastly, in \Cref{fig:model-quality} we highlight our model's predictive performance in various drone configurations.

\subsubsection{Policy Optimisation Ablation Study}
We wanted to evaluate the effect of two design choices: 1) use of model derivatives and 2) use of a critic.
Thus, we implemented the described algorithm once without backpropagating through the model to optimise the actor and critic, instead using the model only as a data source, and once without a value function, instead relying on Monte Carlo estimates of the expected reward.
However, even in the simpler scenario without yaw and using full state observation data as in the \textit{MocapVel} configuration, we were unable to achieve any success with either of these modifications.
Thus, we can only qualitatively, but not quantitatively motivate our algorithmic choices.

\section{Related Work}
\subsection{Quadrotors, Drones \& UAVs}
The potential of classical reinforcement learning methods has already been demonstrated by learning autonomous helicopter flight~\cite{bagnell2001autonomous, ng2006autonomous, abbeel2007application}. 
Later on, deep learning methods enabled further applications. 
In particular, neural networks helped enhance and support classical PID controllers.
The Neural Lander~\cite{shi2019neural} achieves stable drone landing by approximating higher level dynamics, such as air disturbances close to the ground, with a neural network. 
\cite{kaufmann2018deep} uses imitation learning to train an onboard CNN to predict waypoints on a drone racing track. 
\cite{loquercio2019deep} extends this work, dealing with a changing track (\eg moving gates the drone has to pass through).
\cite{li2017learning} has a quadrotor following a person based on a camera input. 
They use model-free reinforcement learning to give goals to a low-level PID controller which translates these into motor commands.

Instead of using a PID controller to steer towards the goal, some approaches use neural networks to predict either thrust-attitude controls or motor commands directly.
Imitation learning was successfully employed in this fashion to train a thrust-attitude controller that is capable of flying through a narrow gap~\cite{lin2019flying}. 
\cite{lambert2019low} replaced the PID controller with MPC in a learnt model using onboard sensors only and hence achieved hovering capability for up to 6 seconds. 
Neuroflight~\cite{koch2019neuroflight} suggested to replace the engineered PID controller of Betaflight\cite{betaflight} with a learnt one that can adapt to the current state of the aircraft. 
Probably closest to us is~\cite{hwangbo2017control} where they also learn a neural network controller that directly computes motor torques based on state inputs. 
They manage to stabilise a quadrotor from random starting conditions even without using an underlying PID controller (for the final policy, they do use one during training/exploration) as in our case. 
However, they achieve this by using a predefined instead of learnt model for simulation. 
Their approach being model-free, they cannot deal with partial observations.

\subsection{Reinforcement Learning}
Model-free deep RL has received a tremendous amount of attention ever since $Q$-learning was successfully applied to playing Atari games directly from raw input images with the use of deep convolutional neural networks~\cite{mnih2015human}.
One key ingredient to its success was its stabilisation of its neural network based $Q$-values using a slowly updating target network.
Work on further stabilising deep $Q$-learning and coping with its other shortcomings like its overestimation bias has since been central to further inquiry~\cite{wang2015dueling, van2016deep, lillicrap2015continuous, fujimoto2018addressing}.
Much of the early deep RL work being limited to discrete action spaces, DDPG \cite{lillicrap2015continuous} revives the idea of deterministic policy gradient and hence, extends $Q$-learning to continuous action spaces. 
A3C \cite{mnih2016asynchronous} takes an Actor-Critic approach and promotes parallelised asynchronous updates. 
Noticeably, the paper also explores n-step value updates, similarly to our value-update schema in \Cref{ssub:value_function}.

Changing focus to \ac{mbrl}, learnt models have been used for planning \cite{chua2018deep, haarnoja2018learning} which have recently closed the gap to model-free methods in simulated physics environments while being more sample efficient.
Looking at gradient-based approaches, Stochastic Value Gradients \cite{heess2015learning} shows how to use a model just for gradient computation of real-world trajectories by backsolving for noise variables.
Model-based value estimation \cite{feinberg2018model} uses short imagined rollouts within the dynamics model to improve the value estimation.
Imagined value gradients \cite{byravan2019imagined} use the dynamics model's gradient to optimise the value function.
Concurrent to our work, \cite{hafner2019dream} demonstrates an optimisation scheme similar to ours on simulated environments.

In robotics applications, end-to-end training of deep visuomotor policies was first shown to be feasible in~\cite{levine2016end}.
\cite{haarnoja2018learning} taught a real-world Minitaur robot to walk in 2 hours using a slightly modified Stochastic Actor Critic method~\cite{haarnoja2018soft}.
\cite{andrychowicz2018learning, akkaya2019solving} learn dexterous in-hand manipulation in a simulator and deploy on a Shadow Dexterous Hand that is robust to various (previously unseen) disturbances.
Closely related to us, \cite{piergiovanni2018learning} optimised a real-world policy by dreaming in a learnt, but deterministic dynamics model.

\subsection{Learning Latent Dynamics of Physical Systems}
\cite{krishnan2017structured, karl2017deep} showed how to learn LSSMs using neural variational inference directly on data sequences using a time-factorised ELBO, the latter pointing out the importance of propagating the reconstruction error through the transition model for improved learning of dynamics.
\cite{fraccaro2017disentangled} showed how to combine neural variational inference with an underlying Kalman smoother in latent space.
Aside from latent variable models, there have also been efforts to inform or restrict deep learning methods using our physical understanding of the world 
\cite{raissi2017physics, de2018end, lutter2019deep}.

\section{Conclusion}
Without encoding any physics knowledge into the latent state-space, we have shown how to learn a drone dynamics model from raw sensor observations.
This model is good enough to learn a controller entirely in simulation that can be executed on a real drone.
The controller is capable of flying the drone to observed goal positions using only onboard sensors and computational resources.
The optimization scheme is based on stochastic analytic gradients enabled by implementing model, actor and critic as differentiable function approximators.
To our knowledge, this is the first \ac{mbrl} approach deployed on a real drone where both the model and controller are learnt by deep learning methods, and one of the first on real hardware.
The methodology was showcased on various drone configurations, but is applicable more broadly to other robotic setups without the need for algorithmic changes.
Thus, we believe that this represents an important step towards learning robots from the ground up using minimal engineering.

However, there are still problems we would like to address in future work, like the reliance on a stabilising PID controller (mainly for initial exploration) or the gap to more acrobatic flight shown in previous works that make use of more prior knowledge.

\appendices

\section{Implementation \& Training}
The model was implemented using TensorFlow~\cite{abadi2016tensorflow} and TensorFlow Probability~\cite{dillon2017tensorflow}.
Training of the models can be done on a single workstation/GPU and takes up to a few days.\\
Mostly the same parameters were used across all experiments and drone configurations.
All components were parametrised by dense neural networks with either 1 or 2 hidden layers with ReLU activations, with the exception of the policy where instead a tanh activation function was used.
The encoder network was chosen to be an RNN in cases where the drone's velocity was unobserved.
We used ADAM\cite{Kingma2014Adam} for optimisation of all our model components.
A list of key hyperparameters is presented in \Cref{tab:hyperparameters}.
\begin{table}[!h]
\renewcommand{\arraystretch}{1.25}
\caption{List of Hyperparameters}
\begin{center}
\begin{tabular}{ll}
\hline 
Parameter & Value \\ 
\hline 
\multicolumn{2}{l}{\textbf{Dynamics Model}} \\ 
batch size & 64\\
dimensions of latents $z_t$ & 32 \\
dimensions of latents $s_t$ & 32 \\
learning rate & \num{3e-4} \\
\# of base matrices & 32 \\
\multicolumn{2}{l}{\textbf{Policy}} \\
batch size & 128 \\
learning rate & \num{1e-4} \\
discount factor & $0.95$\\
\multicolumn{2}{l}{\textbf{Value Function}} \\
batch size & 128\\
learning rate & \num{3e-4} \\
target network learning rate $\alpha_{\phi'}$ & \num{1e-3}\\
\hline 
\end{tabular}
\label{tab:hyperparameters}
\end{center}
\end{table}

\section{Onboard Computational Cost}
For online control of the robot, we do not require execution of all parts of the model. 
In particular, we only need the policy $\policy$ and the filter consisting of inverse measurement model $\qmeas$ and transition models $\ztran$ and $\stran$.
Not needed are the likelihood model for observations $\p[\xi]{x_t}{z_t}$ and rewards $\apxreward$, and the value function as they are only required during training.
We are therefore free to choose the parameterisation of the latter while the former need to be of limited complexity so that they can be executed in real-time on a Raspberry Pi 4.

\section*{Acknowledgment}
The authors would like to thank Djalel Benbouzid for valuable discussions and feedback.

\ifCLASSOPTIONcaptionsoff
  \newpage
\fi

\bibliographystyle{IEEEtran}
\bibliography{main}

\begin{IEEEbiography}[{\includegraphics[width=1in,height=1.25in,clip,keepaspectratio]{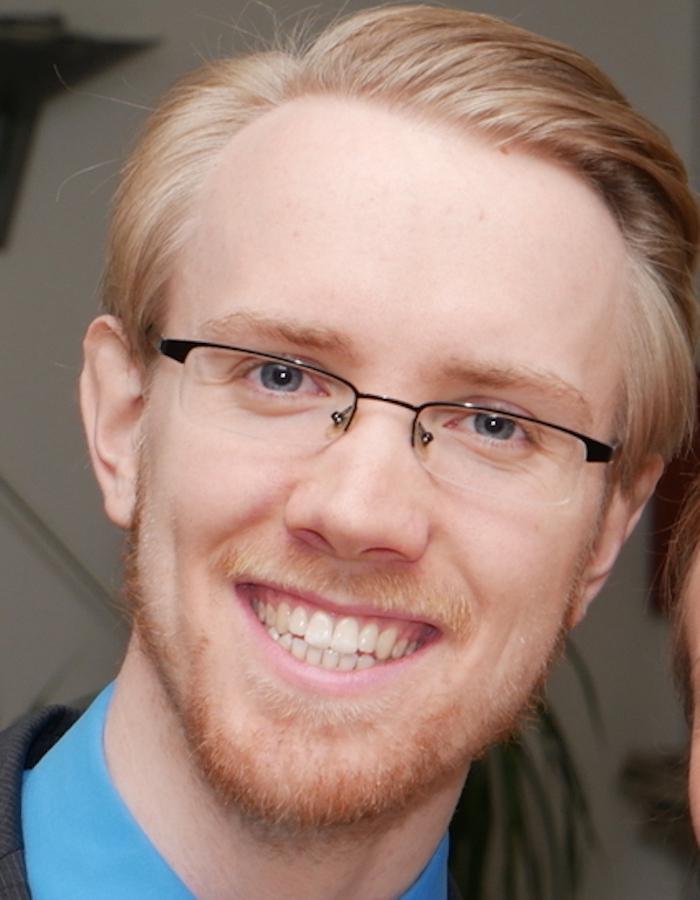}}]{Philip Becker-Ehmck} 
was born in Munich, Germany, in 1992. He received his BSc and MSc degree in computer science from Technical University of Munich in 2014 and 2017, respectively.
He is currently a Ph.D. candidate at Volkswagen Group Machine Learning Research Lab and at the Intelligent Autonomous Systems group of the Technical University of Darmstadt. 
His work focuses on variational inference methods for time series modelling, intrinsic motivation and model-based reinforcement learning.
\end{IEEEbiography}

\begin{IEEEbiography}[{\includegraphics[width=1in,height=1.25in,clip,keepaspectratio]{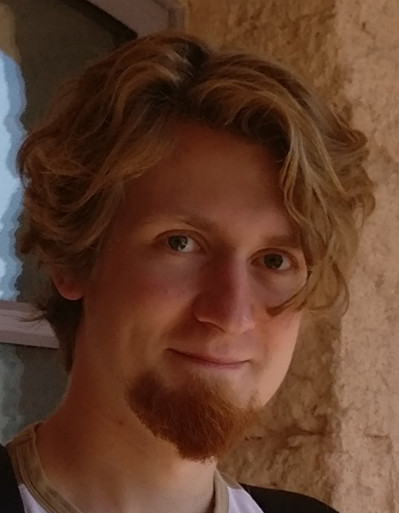}}]{Maximilian Karl}
received his Masters degree in Robotics, Cognition, Intelligence from the Technical University of Munich.
He then started his PhD at the Technical University of Munich in 2014.
In 2017 he joined the Volkswagen Group Machine Learning Research Lab as a research scientist.
In his thesis he focuses on modelling robotic environments with latent state-space models and controlling them with intrinsic motivation, also known as unsupervised control.
His research interests include intrinsic motivation, variational inference, stochastic optimal control, reinforcement learning and information theory.
\end{IEEEbiography}

\begin{IEEEbiography}[{\includegraphics[width=1in,height=1.25in,clip,keepaspectratio]{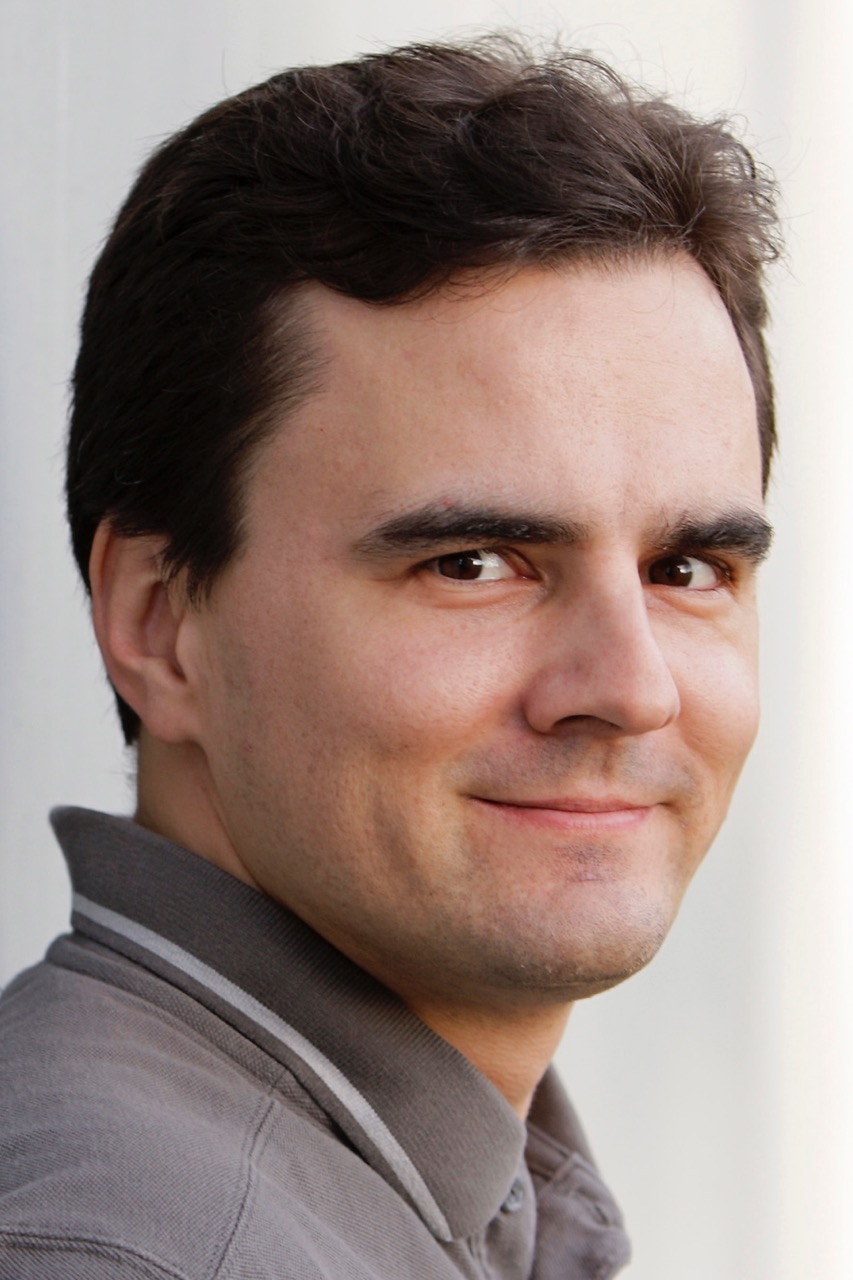}}]{Jan Peters}
is a professor for Intelligent Autonomous Systems at the Computer Science Department of the Technical University of Darmstadt, and at the same time a senior research scientist and group leader at the Max-Planck Institute for Intelligent Systems, where he heads the interdepartmental Robot Learning Group. 
Jan Peters has received the  Dick Volz Best 2007 US Ph.D. Thesis Runner-Up Award, the Robotics: Science \& Systems -- Early Career Spotlight, the INNS Young Investigator Award, and the IEEE Robotics \& Automation Society's Early Career Award as well as numerous best  paper awards. 
In 2015, he received an ERC Starting Grant and in 2019, he was appointed as an IEEE Fellow.
\end{IEEEbiography}

\begin{IEEEbiography}[{\includegraphics[width=1in,height=1.25in,clip,keepaspectratio]{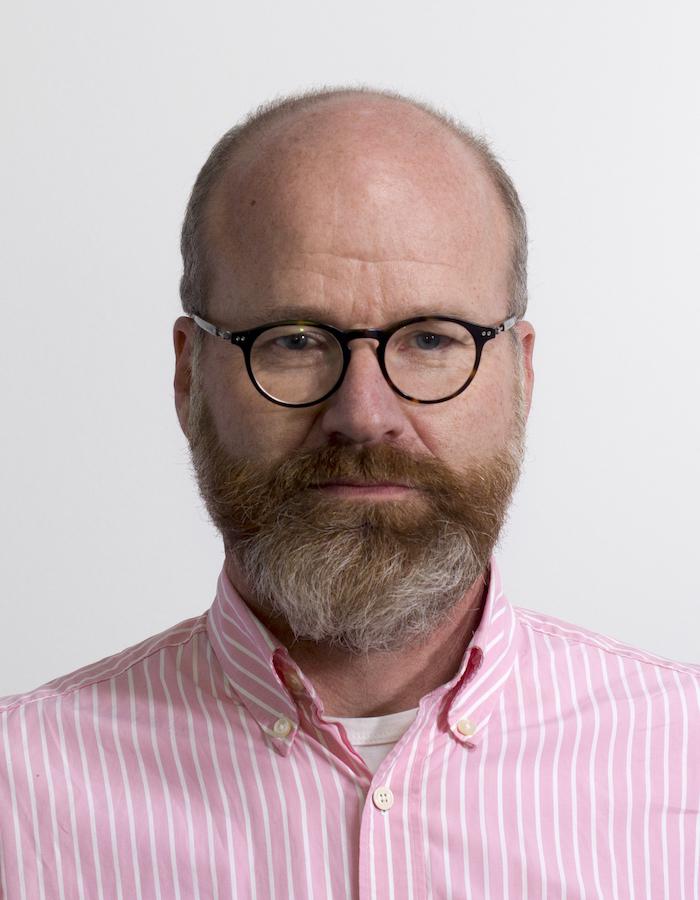}}]{Patrick van der Smagt}
received his Ph.D. degree in mathematics and computer science from the University of Amsterdam. 
He is director of AI Research at Volkswagen Group, head of the Volkswagen Group Machine Learning Research Lab in Munich, and holds a honorary professorship at ELTE University Budapest. 
He previously directed a lab as professor for machine learning and biomimetic robotics at the Technical University of Munich while leading the machine learning group at the research institute fortiss.
Patrick van der Smagt has won numerous awards, including the 2013 Helmholtz-Association Erwin Schr{\"o}dinger Award, the 2014 King-Sun Fu Memorial Award, the 2013 Harvard Medical School/MGH Martin Research Prize, the 2018 Webit Best Implementation of AI Award, and best-paper awards at machine learning and robotics conferences and journals.
\end{IEEEbiography}

\end{document}